\documentclass[12pt, draftclsnofoot, onecolumn]{IEEEtran}
\usepackage{times}
\usepackage[final]{graphicx}
\usepackage{textcomp}
\usepackage[reqno]{amsmath}
\usepackage{amsfonts}
\usepackage{times,amsmath,epsfig}
\usepackage{latexsym,amssymb}
\usepackage{cite}
\usepackage{psfrag}
\usepackage{stfloats}
\usepackage{float}
\usepackage{amsmath}
\usepackage{amssymb}
\usepackage{array}
\usepackage[noend]{algorithmic}
\usepackage[ruled, linesnumbered]{algorithm2e}
\usepackage{bm}
\usepackage{color}
\usepackage{multicol}
\usepackage{multirow}
\usepackage{setspace}
\usepackage{tabularx}
\usepackage[T1]{fontenc}
\usepackage{subeqnarray}
\usepackage{cases}
\usepackage{makecell}

\newtheorem{lem}{Lemma}
\newtheorem{prob}{Problem}
\newtheorem{remark}{\bf{Remark}}
\usepackage{subfig}
\usepackage{pdfpages}
\usepackage{graphicx}
\newlength{\figwidth}
\newcommand{\PreserveBackslash}[1]{\let \temp =\\#1 \let \\ = \temp}
\newcolumntype{C}[1]{>{\PreserveBackslash\centering}p{#1}}
\newcolumntype{R}[1]{>{\PreserveBackslash\raggedleft}p{#1}}
\newcolumntype{L}[1]{>{\PreserveBackslash\raggedright}p{#1}}

\begin{document}

\title{Joint Power Control and Data Size Selection for Over-the-Air Computation Aided Federated Learning}

\author{
Xuming An, Rongfei Fan, Shiyuan Zuo, Han Hu, Hai Jiang, and Ning Zhang
\thanks
{
X. An, S. Zuo, and H. Hu are with the School of Information and Electronics, Beijing Institute of Technology, Beijing 100081, P. R. China. (\{3120195381,3120210836,hhu\}@bit.edu.cn).
}
\thanks{R.~Fan is with the School of Cyberspace Science and Technology, Beijing 100081, P. R. China. (fanrongfei@bit.edu.cn).}
\thanks{H.~Jiang is with the Department of Electrical and Computer Engineering, University of Alberta, Edmonton, AB T6G 1H9, Canada (hai1@ualberta.ca).}
\thanks{N.~Zhang is with the Department of Electrical and Computer Engineering, University of Windsor, Windsor, ON, N9B 3P4, Canada (ning.zhang@uwindsor.ca).}
}

\maketitle

\begin{abstract}
Federated learning (FL) has emerged as an appealing machine learning approach to deal with massive raw data generated at multiple mobile devices, {which needs to aggregate the training model parameter of every mobile device at one base station (BS) iteratively}. For parameter aggregating in FL, over-the-air computation is a spectrum-efficient solution, which allows all mobile devices to transmit their parameter-mapped signals concurrently to a BS. Due to heterogeneous channel fading and noise, there exists difference between the BS's received signal and its desired signal, measured as the mean-squared error (MSE). To minimize the MSE, we propose to jointly optimize the signal amplification factors at the BS and the mobile devices as well as the data size (the number of data samples involved in local training) at every mobile device.
{The formulated problem is challenging to solve due to its non-convexity.
To find the optimal solution, with some simplification on cost function and variable replacement, which still preserves equivalence, we transform the changed problem to be a bi-level problem equivalently. }
{For the lower-level problem, optimal solution is found by enumerating every candidate solution from the Karush-Kuhn-Tucker (KKT) condition.}
For the upper-level problem,  {the} optimal solution is found by exploring its piecewise convexity.
{
Numerical results show that our proposed method can greatly reduce the MSE and can help to improve the training performance of FL compared with benchmark methods.}
\end{abstract}

\begin{IEEEkeywords}
Over-the-air computation, federated learning, power control, data size selection.
\end{IEEEkeywords}

\section{Introduction}\label{sec:intro}

In recent years, mobile devices, including smartphones and sensors, have experienced an exponential growth\cite{xu2020client,zhu2019broadband}. A massive amount of data are generated from these mobile devices and promote various kinds of machine learning based applications, such as disaster warning based on digital twin, health monitoring via wearable devices, user habit learning through activities of smartphone holders, etc. \cite{goodfellow2016deep}.
Traditionally, machine learning approaches train a group of labeled data in a centralized way \cite{jeon2020compressive}. So mobile devices need to upload their local raw data to a {central server}, which may consume a large amount of wireless communication resources and cause privacy {issues}.

To tackle the above {issues}, a new distributed model training architecture is proposed, called {\it Federated Learning (FL)} \cite{conf/aistats/McMahanMRHA17,ahn2019wireless,wang2019adaptive,conf/iclr/YangFL21}. In an FL system, multiple mobile devices can collaboratively learn a common model under the coordination of a base station (BS) by iteratively exchanging model parameters between the BS and the mobile devices \cite{wang2019edge}.
{In each iteration, the model parameters are updated separately by every mobile device, based on the common model parameter broadcasted by the BS in last iteration and the dataset at local.
Then every mobile device uploads the information of its updated model parameters to the BS, who afterwards fuses the received information and broadcasts the aggregated model parameter to every mobile device.}
In this process, the private raw data of each mobile device is not shared with the BS, and thus, the privacy {is protected at some extent}.
When the BS performs data fusion,  it takes a weighted sum of the information received from the mobile devices. The weight associated with a mobile device depends on its data size involved in its local training.

In FL, iterations of data exchange involve frequent multiple-access communications between mobile devices and the BS, which {is} spectrum- and/or energy-consuming, especially when there are a lot of mobile devices or the spectrum resources are limited. To overcome this problem, over-the-air computation can be adopted, which is able to directly calculate the summation of uploaded data that are transmitted simultaneously from multiple mobile devices to the BS \cite{nazer2007computation,yang2020federated}, thanks to the signal supposition property of the multiple-access channel.
Over-the-air computation can achieve efficient data fusion, since uploading from multiple mobile devices happens simultaneously, {and has shown great potential for not only FL but also  wireless sensor network \cite{liu2020over}}.

With general over-the-air computation (which may or may not work with FL), since the mobile devices' transmitted signals experience heterogenous channel fading, the superposed signal cannot be exactly identical to the desired one. Hence mean-squared error (MSE) of the aggregated signal at the BS is usually taken as the performance metric.
{For an over-the-air computation system supporting FL, the MSE is also proved to be highly related to the training loss\cite{cao2021optimized,journals/jsac/CaoZXC22,wang2021federated}, which has also been disclosed in Section \ref{s:convergence}.
}
To combat the heterogeneous channel fading and reduce the MSE, each mobile device amplifies its signal to be transmitted, and the BS also amplifies its received signal.
The amplification factor at every mobile device and the BS are usually optimized jointly to minimize the MSE \cite{liu2020over,cao2020optimized}. It is always desired that the MSE should be reduced as much as possible.

When FL is {conducted} by over-the-air computation, recalling that the weight associated with a mobile device depends on its {data size} involved in its local training, it can be seen that the MSE is a function of these mobile devices'  {data size}.
{On the other hand, due to the existence of overfitting, it may not be necessary to use up all the available data samples for a general machine learning task {\cite{books/Bishop07}}. FL is not an exception either {\cite{journals/ftml/KairouzMABBBBCC21}}.
Besides, using less data samples for local training can also help to save computation burden for every mobile device.
Some literature on FL have already followed this idea  \cite{conf/icml/YuJY19,conf/nips/HaddadpourKMC19,conf/icml/MurataS21,conf/aistats/HaddadpourKMM21,conf/icml/AvdiukhinK21,conf/nips/ZhuLLLH21}.
Hence we can adjust the data size of each mobile device for local training.
}

 {Based on the above discussion, we can expect that the MSE can be further reduced by adjusting the data size} of every mobile device in a proper way.
Motivated by this, in this paper, we investigate an FL system supported by the over-the-air computation technique, and we jointly optimize each mobile device's amplification factor and {data size} and the BS's amplification factor, so as to minimize the MSE.

\subsection {Main Contributions}

The main contributions of this paper are summarized as follows:

\begin{itemize}
\item \textbf{A new perspective to reduce MSE:} For an FL system supported by over-the-air computation, in addition to adjusting the amplification factors at the mobile devices and the BS, we propose to also adjust the  {data} size of every mobile device in the local training stage, so as to further reduce the MSE. An optimization problem is formulated.
{
\item \textbf{Problem transformation:} In the formulated problem, the cost function is with the amplification factors of the BS and mobile devices, denoted as $a$ and $b_k$ for $k$th mobile device, and the data size of the mobile devices, denoted as $S_k$ for $k$th mobile device, coupled. The cost function also involves the indicator function of $S_k$.
Either the coupling among $a$, $\{b_k\}$, and $\{S_k\}$ or the existence of indicator function with $\{S_k\}$ makes the cost function non-convex with the data size $\{S_k\}$. To solve this challenge, we remove the indicator function from the cost function with proved equivalence and transform variable $S_k$ to another variable $\beta_k$. With these operations, the cost function of the transformed problem is a convex function of $\{\beta_k\}$, which paves the way for the optimal solving of an equivalently transformed problem in the sequel.}
{
\item \textbf{Two-level structure to solve the transformed problem}. The transformed problem is {still} not jointly convex with {$a$, $\{b_k\}$, and $\{\beta_k\}$  because of the coupling among them}. We propose to decouple them and decompose the problem into two levels. In the lower-level problem, the BS's amplification factor $a$ is given, and we optimize the mobile devices' amplification factors $\{b_k\}$ and variables $\{\beta_k\}$. In the upper-level problem, the BS's amplification factor $a$ is optimized. Through this operation, the transformed problem is equivalent with the upper-level problem.}
    {
\item \textbf{Deriving closed-form solution for the lower-level problem:} In the lower-level problem, the cost function is jointly convex with respect to the mobile devices' amplification factors $\{b_k\}$ and variables $\{\beta_k\}$. To find the optimal solution of the lower-level problem, we investigates the Karush-Kuhn-Tucker (KKT) condition of the lower-level problem and derive closed-form solution through exhaustively exploring the candidate solutions.}
\item \textbf{Solving the upper-level problem with unclear convexity:} For the upper-level problem, the convexity is unclear and hard to explore. To overcome this challenge, {we discover the implicit convexity of the cost function over multiple intervals via the derived closed-form solution for the lower-level problem and show how to characterize these intervals.}
Accordingly, we use convex optimization techniques to find the optimal solution (i.e., the minimal MSE) in each interval, and the optimal solution of the upper-level problem can be found by comparing the minimal MSE values in all the intervals.
\end{itemize}

\subsection{Related Work}\label{sec:related}

In general FL research (which does not adopt over-the-air computation), two major topics are on {the analysis of convergence to optimal/stationary solution and the minimization of time delay to convergence.}
On convergence analysis, with some assumptions on the convexity and smoothness of the loss function for training, the gap to the global optimal loss function is  analyzed versus the number of iterations, which is expected to converge to zero with a high convergence rate, under various configuration of step size and searching gradient \cite{chen2021AJoint,shi2020joint,chen2020convergence,yang2020energy,9497677,zhang2021adaptive,conf/icml/YuJY19,conf/nips/HaddadpourKMC19,conf/icml/MurataS21,luo2020hfel,vu2020cell}.
{Specially, \cite{conf/icml/YuJY19,conf/nips/HaddadpourKMC19,conf/icml/MurataS21} assume the data size of each mobile device for local computing is adjustable.}
{
On time delay minimization, \cite{luo2020hfel} proposes a hierarchical FL framework where there is one cloud server connecting multiple BS's. The association between BS and mobile devices, together with mobile devices' CPU frequency and allocated bandwidth for data uploading, are jointly optimized.
\cite{vu2020cell} optimizes mobile devices' CPU frequency, downloading and uploading date rate, and transmit power for a cell-free massive MIMO network.}

For over-the-air computation research, MSE is {an} important metric and heterogeneous channel fading between mobile devices and the BS is the major cause to increase the MSE.
To overcome the heterogenous channel fading,
\cite{liu2020over} and \cite{cao2020optimized} are two pioneering works, in which the amplification factors at mobile devices and the BS are jointly optimized to minimize the MSE.
In \cite{zhai2021hybrid}, the authors adopt multiple antennas to combat the channel fading between the mobile devices and the BS, and digital beamforming at every mobile device, together with the hybrid beamforming at the BS, are performed jointly to reduce the MSE.

Recently there have been some research efforts on FL systems supported by over-the-air computation technique.
In \cite{zhu2019broadband}, a multiple-band system with every band supporting the over-the-air computation is investigated. Time delay for one round of data aggregation under this setup is analyzed and is shown to outperform the system using the traditional digital orthogonal frequency-division multiplexing (OFDM) technique.
In \cite{sery2021over}, convergence analysis is performed when there is no power control. This work shows that the FL with an over-the-air computation technique can achieve the optimal solution, when the number of enrolling mobile devices is infinite.
The work in \cite{cao2021optimized} performs a convergence analysis when there is power control at every mobile device. It is shown that FL with over-the-air computation technique can converge to the optimal solution, when the number of enrolling mobile devices is finite, and the converging speed is highly related to the MSE.
In \cite{xu2021learning}, private learning rate of every mobile device is set to combat the distortion caused by heterogeneous channel fading between every mobile device and the BS. To minimize the MSE, every mobile device's learning rate is optimized dynamically over aggregation round in multiple-input-single-output (MISO) and multiple-input-multiple-output (MIMO) scenarios.
In \cite{sun2021dynamic}, with an energy consumption budget imposed on every mobile device, dynamic scheduling (which determines the group of nodes that can access the radio channel) in each round of aggregation is investigated to minimize the loss function.
In \cite{liu2021reconfigurable,wang2021federated}, a system aided by reconfigurable intelligent surface (RIS) is considered. The selection of the mobile devices, beamforming at the BS, and the phase shift of RIS's every element in each round of iteration are optimized jointly to promote convergence by reducing the optimality gap \cite{liu2021reconfigurable} or by minimizing the MSE \cite{wang2021federated}.

In summary, MSE is critical in FL systems supported by over-the-air computation technique, and in the literature, there have been research efforts on MSE reduction/minimization, by, for example, adjusting the amplification factors at the mobile devices and the BS. Different from the literature, {especially \cite{liu2020over} and \cite{cao2020optimized}}, we propose a new dimension to minimize the MSE, i.e., in addition to adjusting the amplification factors, we propose to also adjust the  {data size} of every mobile device in local training stage, so as to further reduce the MSE.
Due to the inclusion of  {data size} optimization, the formulated joint optimization problem is much harder to solve than problems with only amplification factor optimization. Accordingly, we develop a new {method} to solve our formulated joint optimization problem.

\subsection {Paper Organization}
The rest of this paper is organized as follows:
Section \ref{s:model} introduces the system model and problem formulation.
Section \ref{s:opt_sol} gives our structure to solve the problem. Sections \ref{s:optimal_solution_beta_b} and \ref{s:optimal_solution_a} demonstrate how the lower-level and upper-level optimization problems are solved, respectively. Section \ref{s:numerical_results} shows numerical results, followed by our conclusion in Section \ref{s:conclusion}.

\section{System Model and Problem Formulation} \label{s:model}

Consider a wireless network with one BS and $K$ wireless-linked mobile devices, which constitute the set $\mathcal{K} \triangleq \{1, 2, ..., K\}$.
Each mobile device collects labelled data points independently and generates its own local data set.
The data points collected by all the mobile devices are supposed to follow the same statistical heterogeneity \cite{chen2020convergence}.
For the $k$th mobile device, we assume the associated data set is $\mathcal{D}_k$ with $D_k$ elements. The $l$th element, i.e., the $l$th labelled data point,  of set $\mathcal{D}_k$ can be written as $\{\bm{x}_{k,l}, {y}_{k,l}\}$ for $l=1, 2, ..., {D}_k$, where $\bm{x}_{k,l} \in \mathcal{R}^{N_{\text{in}}}$ is the input vector and ${y}_{k,l} \in \mathcal{R}$ is the output scaler.
To perform data analysis, with the collected data points from these $K$ mobile devices, the BS needs to train a parameter vector $\bm{w}\in \mathcal{R}^{N_{\text{in}}}$ (referred to as  the training problem) to minimize $\frac{1}{ \sum_{k=1}^{K} D_k }  \sum_{k=1}^{K} \sum_{l=1}^{D_k}f\left(\bm{w}, \bm{x}_{k,l}, {y}_{k,l} \right)$,
where $f\left(\bm{w}, \bm{x}_{k,l}, {y}_{k,l} \right)$ is the loss function to evaluate the error for approximating the output value ${y}_{k,l}$ by the input vector $\bm{x}_{k,l}$ under a selection of parameter $\bm{w}$.
FL is implemented in a distributed manner. It has a number of iterations. Each iteration has three rounds as follows.
\begin{itemize}
\item Round 1: With the most-updated parameter vector $\bm{w}$ broadcasted from the BS to every mobile device, every mobile device utilizes its local data set to generate a local gradient vector of the parameter vector; 
\item Round 2: Every mobile device uploads its local gradient vector of parameter vector $\bm{w}$ to the BS;
\item Round 3: The BS {aggregates} the gradient vectors from every mobile device to generate a new parameter vector $\bm{w}$.
\end{itemize}
The above procedure is repeated until the parameter vector converges or there are a sufficient number of iterations.

Take the $t$th iteration as an example. In Round 1, {to save computation burden and reduce the MSE for gradient vector aggregation (to be explained in the sequel)}, not all the local data points are utilized {for local updating}.

{Suppose a number of $S_k$ data points are selected from the $k$th mobile device's local dataset for  local updating, which is also called as data size of $k$th mobile device.
Then there is $S_k \leq D_k, \forall k\in \mathcal{K}$.
}
Further, {to promise the performance of FL, the total number of data points involved in training should be no less than a threshold $S_T$ ($S_T \leq \sum_{k=1}^{K} D_k$), i.e.,
$\sum_{k=1}^{K} S_k \geq S_T.$

In Round 2, the technique of over-the-air computation is adopted, which is efficient when the receiver wants to collect the weighted sum of multiple transmitters. Suppose the gradient to be transmitted by the $k$th mobile device is $\bm{\nabla}_k = \frac{\partial f_k (\bm{w}_t, S_k)}{\partial \bm{w}_t}$, where $\bm{w}_t$ is the broadcasted parameter vector $\bm{w}$ in the $t$th iteration, and the function $f_k(\bm{w}_t, S_k)$ is defined as
{
$f_k(\bm{w}_t, S_k) \triangleq \frac{1}{ S_k }   \sum_{l=1}^{S_k}f\left(\bm{w}_t, \bm{x}_{k, \zeta_l }, {y}_{k,  \zeta_l} \right)$, where $\zeta_l$ represents the index of $l$th selected data points.
}
Suppose the channel coefficient between the $k$th mobile device and the BS is $h_k$, {which is a real number} \footnote{{This assumption is made by following \cite{cao2020optimized}, considering that the phase shift due to channel fading can be compensated by performing channel estimation. The channel estimation can be done by broadcasting downlink pilots from the BS on selected frequency band, thanks to the uplink-downlink channel reciprocity. }},
and the signal amplification factor of the $k$th mobile device is $b_k$, for $k\in \mathcal{K}$.
Denote the amplification factor of the received signal at the BS as $a$.
Then with the technique of over-the-air computation, the recovered signal at the BS is
\begin{equation}
\hat{\bm{z}}_t = a \sum_{k=1}^{K}  b_k h_k \cdot \bm{{\nabla}}_k + a \cdot \bm{n}_t
\end{equation}
where $\bm{n}_t$ is the noise vector with every element following a Gaussian distribution $\mathcal{N}(0, \sigma^2)$.
To promise data diversity, it needs to be satisfied that $b_k >0$ for $k \in \mathcal{N}$.
It should be also noticed that the capability of amplifying the transmitted signal at every mobile device is limited. Then there is an upper bound of $b_k$, which is denoted as $b_k^{\max}$ for $k \in \mathcal{K}$.
On the other hand, there {is} no limitation on $a$ at the BS. The reason is as follows. After receiving the aggregated signal at the BS, the BS can first digitalize the received signal through quantization. Then the BS can amplify the digitalized signal at any ratio.
Due to the noise signal $\bm{n}_t$ and the heterogeneous channel coefficients $h_k$'s among multiple mobile devices, there is a distortion between the recovered signal $\hat{\bm{z}}_t$ and the ideal aggregated gradient $\bm{z}_t$, which can be written as
\begin{equation}
\bm{z}_t =  \sum_{k=1}^{K} \left(\frac{S_k}{\sum_{k=1}^{K} S_k}\right)
   \bm{\nabla}_k.
\end{equation}
The metric of MSE is usually adopted to measure the distortion between $\hat{\bm{z}}_t$ and $\bm{z}_t$, which can be given as
\begin{equation} \label{e:MSE_raw}
\begin{array}{ll}
   \text{MSE}&= \mathbb{E} \{||\hat{\bm{z}}_t - \bm{z}_t||^2\} \\
             &= \sum_{k=1}^{K}  \left( a  b_k h_k  -  \left(\frac{S_k}{\sum_{k=1}^{K} S_k}\right) \right)^2
             \mathbb{E} \{||\{ \bm{\nabla}_k\}||^2 \} {\mathcal{I}\left(S_k >0 \right)} + a^2 \sigma^2 \\
             &= \sum_{k=1}^{K}  \left( a  b_k h_k  -  \left(\frac{S_k}{\sum_{k=1}^{K} S_k}\right) \right)^2 {c_k \mathcal{I}\left(S_k >0 \right)}  + a^2 \sigma^2
\end{array}
\end{equation}
with $\mathbb{E}\{\cdot\}$ denoting expectation, {$\mathbb{E} \{||\bm{{\nabla}}_k||^2\} $ defined as $c_k$ for $k \in \mathcal{K}$ for the ease of presentation in the following, and $\mathcal{I}(\cdot)$ defined as the indicator function.}
{
For the defined MSE in (\ref{e:MSE_raw}), the $\mathcal{I}(S_k)$ would be zero if $S_k =0$ and would be one otherwise. This is because when $S_k=0$, the $k$th mobile device actually does not take part in the gradient aggregation, and the associated MSE will not count the gradient aggregation distortion lead by $k$th mobile device.
}

In Round 3 for the $t$th iteration of FL procedure, the BS utilizes the recovered signal $\hat{\bm{z}}_t$, which represents an approximation of $\bm{z}_t$, to update the parameter vector $\bm{w}$ as follows
\begin{equation}
\bm{w}_{t+1} = \bm{w}_t - \eta \hat{\bm{z}}_{t}
\end{equation}
where $\eta$ is a pre-defined step-size.

{
In the procedure of FL with over-the-air computation, the MSE as expressed in (\ref{e:MSE_raw}), which represents the distortion between $\hat{\bm{z}}_t$ and $\bm{z}_t$, has been proved to be highly related with the training loss\cite{cao2021optimized,journals/jsac/CaoZXC22,wang2021federated}, which has also been disclosed in Section \ref{s:convergence}. To suppress the training loss, as expected in every training task, the MSE should be kept as small as possible.}
Thus, in this paper, our target is to minimize the MSE, through optimizing the variables $b_k$, $S_k$ for  $k \in \mathcal{K}$ and $a$.
Accordingly, the following optimization problem is formulated.
\begin{prob} \label{p:raw_opt_orig}
\begin{subequations}
\begin{align}
\mathop{\min} \limits_{a, \{b_k\}, \{S_k\}} & \quad \sum_{k=1}^{K}  \left(a b_k h_k  - \left(\frac{S_k}{\sum_{k=1}^{K} S_k}\right) \right)^2 { c_k {\mathcal{I}\left(S_k >0 \right)} } + a^2 \sigma^2 \nonumber \\
\text{s.t.}   & \quad 0 {\leq} b_k \leq b_k^{\max}, \forall k \in \mathcal{K}, \label{e:raw_opt_orig_b}\\
                         & \quad 0 {\leq} S_k \leq D_k, \forall k \in \mathcal{K}, \label{e:S_k_interval}\\
                         & \quad \sum_{k=1}^K S_k \geq S_T, \label{e:raw_opt_orig_S_k_sum}\\
                         & \quad a>0.
\end{align}
\end{subequations}
\end{prob}

\section{Optimal Solution Structure} \label{s:opt_sol}

{
Looking into Problem \ref{p:raw_opt_orig}'s objective function, it is the summation of the term $a^2 \sigma^2$ and $\left(a b_k h_k  - \left(\frac{S_k}{\sum_{k=1}^{K} S_k}\right) \right)^2 { c_k {\mathcal{I}\left(S_k >0 \right)} }$ for $k\in \mathcal{K}$, which are always no less than zero.
Intuitively, the cost function will achieve its minimal value $a^2 \sigma^2$ if
$\{b_k\}$ and $\{S_k\}$ are properly selected to make the term $\left(a b_k h_k  - \left(\frac{S_k}{\sum_{k=1}^{K} S_k}\right) \right)^2 { c_k {\mathcal{I}\left(S_k >0 \right)} }$ to be zero for every $ k \in \mathcal{K}$, especially considering that $b_k$ can be set separately to make $a b_k h_k$ offset $\frac{S_k}{\sum_{k=1}^K S_k}$ and $S_k$ can be set as 0 even this offset cannot be achieved for $k$.
However, due to the existence of upper bound of $b_k$ defined in (\ref{e:raw_opt_orig_b}), we cannot promise to make the above offset happen for every $k\in \mathcal{K}$.
Moreover, we cannot always require $S_k$ to be zero for every $k$ such that the above offset does not succeed, because of the constraint (\ref{e:raw_opt_orig_S_k_sum}) imposed on $\{S_k\}$.
In one word, Problem \ref{p:raw_opt_orig}'s objective function cannot achieve its lower bound $a^2 \sigma^2$ easily.
}

{What is even worse,
in the cost function of Problem \ref{p:raw_opt_orig},
it can be also observed that:
}
1) there exists indicator function $\mathcal{I}(S_k>0)$ for $k\in \mathcal{K}$,
2) the set of $\{S_k\}$ are coupled each other in the term $\frac{S_k}{\sum_{k=1}^K S_k}$ for $k\in \mathcal{K}$,
3) $a$ and $b_k$ are entangled for $k\in \mathcal{K}$.
For the above three listed observations, any one of them can lead to the non-convexity of Problem \ref{p:raw_opt_orig}, which brings challenge into problem solving.
In this sequel, we will show how to find the optimal solution of Problem \ref{p:raw_opt_orig} through
{
simplification, transformation, decomposition, and subsequent analysis on the decomposed problems.
}

{
\subsection{Simplification of Cost Function}
For Problem \ref{p:raw_opt_orig}, the following lemma can be expected, which can help to remove the indicator function $\mathcal{I}(S_k)$ for $k\in \mathcal{K}$ from its cost function while still preserving equivalence.
\begin{lem} \label{lem:indicator_remove}
Problem \ref{p:raw_opt_orig} is equivalent with Problem \ref{p:raw_opt}, which is given as
\begin{prob} \label{p:raw_opt}
\begin{subequations}
\begin{align}
\mathop{\min} \limits_{a, \{b_k\}, \{S_k\}} & \quad \sum_{k=1}^{K}  \left(a b_k h_k  - \left(\frac{S_k}{\sum_{k=1}^{K} S_k}\right) \right)^2 {c_k }+ a^2 \sigma^2 \nonumber \\
\text{s.t.}   & \quad 0 \leq b_k \leq b_k^{\max}, \forall k \in \mathcal{K}, \\
                         & \quad 0 \leq S_k \leq D_k, \forall k \in \mathcal{K}, \label{e:S_k_interval}\\
                         & \quad \sum_{k=1}^K S_k \geq S_T, \label{e:S_k_sum}\\
                         & \quad a>0.
\end{align}
\end{subequations}
\end{prob}
\end{lem}
}

\begin{IEEEproof}

Please refer to Appendix \ref{app:indicator_remove}.

\end{IEEEproof}

{
\subsection{Transformation of Variables}
With the equivalence between Problem \ref{p:raw_opt_orig} and Problem \ref{p:raw_opt} disclosed in Lemma \ref{lem:indicator_remove}, we only need to solve Problem \ref{p:raw_opt}. In this subsection, the set of variables $\{S_k\}$ in the cost function of Problem \ref{p:raw_opt} will be decoupled.
}
Define
\begin{equation} \label{e:beta_def}
\beta_k = \frac{S_k}{ \sum_{k=1}^{K} S_k},  \forall k \in \mathcal{K},
\end{equation}
and define
\begin{equation} \label{e:Xi_def}
\Xi = \frac{1}{\sum_{k=1}^K S_k},
\end{equation}
which can be found to lie in the interval $\left[\frac{1}{ \sum_{k=1}^K D_k }, \frac{1}{S_T}\right]$. Then we have
\begin{equation} \label{e:S_k_by_Xi}
S_k = \frac{\beta_k}{\Xi}, \forall k\in \mathcal{K}.
\end{equation}
With the constraint of
$S_k$ for $k\in \mathcal{K}$ in (\ref{e:S_k_interval}), we have
\begin{equation}
0 {\leq} \frac{\beta_k}{\Xi} \leq D_k, \forall k \in \mathcal{K}.
\end{equation}
Considering the fact that $\Xi \in \left[\frac{1}{ \sum_{k=1}^K D_k }, \frac{1}{S_T}\right]$, it can be derived that
\begin{equation} \label{e:box_beta_0}
0 {\leq} \beta_k \leq {D_k} \times \Xi \leq \frac{D_k}{S_T} \triangleq \beta_k^{\max}, \forall k \in \mathcal{K},
\end{equation}
which can be simplified as
\begin{equation} \label{e:box_beta}
0 {\leq} \beta_k \leq  \beta_k^{\max}, \forall k \in \mathcal{K}.
\end{equation}
Additionally, according to the definition of $\beta_k$ for $k\in \mathcal{K}$, the set of $\beta_k$ for $k\in \mathcal{K}$ should satisfy
\begin{equation} \label{e:sum_beta}
\sum_{k=1}^K \beta_k = 1.
\end{equation}

For the transformation from the set of variables $\{S_k\}$ to the set of variables $\{\beta_k\}$, the following lemma can be anticipated.
\begin{lem} \label{lem:S2beta}
There is one-to-one mapping between the feasible region of $\{S_k\}$ defined by (\ref{e:S_k_interval}) and (\ref{e:S_k_sum}) and the feasible region of $\{\beta_k\}$ defined by (\ref{e:box_beta}) and (\ref{e:sum_beta}).
\end{lem}
\begin{IEEEproof}
Please refer to the proof in Appendix \ref{app:S2beta}.
\end{IEEEproof}

{According to Lemma \ref{lem:S2beta}}, we can use $\{\beta_k\}$ instead of $\{S_k\}$ in Problem
\ref{p:raw_opt}.
Accordingly, Problem \ref{p:raw_opt} is equivalent to the following optimization problem
\begin{prob} \label{p:frac_opt}
\begin{subequations}
\begin{align}
\mathop{\min} \limits_{a, \{b_k\}, \{\beta_k\}} & \quad \sum_{k=1}^{K}  \left(a b_k h_k  -\beta_k \right)^2 {c_k}+ a^2 \sigma^2 \nonumber \\
\text{s.t.}   & \quad 0 {\leq} b_k \leq b_k^{\max}, \forall k \in \mathcal{K}, \\
                         & \quad 0 {\leq} \beta_k \leq \beta_k^{\max}, \forall k \in \mathcal{K},\\
                         & \quad \sum_{k=1}^K \beta_k = 1, \\
                         & \quad a>0.
\end{align}
\end{subequations}
\end{prob}

\subsection{Decomposition}
The cost function of Problem \ref{p:frac_opt} is a convex function of $\{\beta_k\}$ and $\{b_k\}$. However, it is not a joint convex function of $a$, $\{b_k\}$ and $\{\beta_k\}$. Thus, Problem \ref{p:frac_opt} is still non-convex. In the following, we will decompose Problem \ref{p:frac_opt} into two levels. In the lower level, with the variable $a$ given, all other variables are optimized, in which case the cost function $E(a)$ is achieved. In the upper level, the variable $a$ is optimized to find the minimal $E(a)$. Specifically, the lower-level optimization problem is given as
\begin{prob} \label{p:lower_level_raw}
\begin{subequations}
\begin{align}
E(a) \triangleq \mathop{\min} \limits_{\{b_k\}, \{\beta_k\}} & \quad \sum_{k=1}^{K} \left(a b_k h_k  - \beta_k\right)^2 {c_k}+ a^2 \sigma^2 \nonumber \\
\text{s.t.}   & \quad 0 {\leq} b_k \leq b_k^{\max}, \forall k \in \mathcal{K}, \\
                         & \quad 0 {\leq} \beta_k \leq \beta_k^{\max}, \forall k \in \mathcal{K},\\
                         & \quad \sum_{k=1}^K \beta_k = 1.
\end{align}
\end{subequations}
\end{prob}

The upper-level optimization problem is given as
\begin{prob} \label{p:upper_level_raw}
\begin{subequations}
\begin{align}
\mathop{\min} \limits_{a} & \quad E(a) \nonumber \\
\text{s.t.}   & \quad a>0.
\end{align}
\end{subequations}
\end{prob}
It can be checked Problem \ref{p:upper_level_raw} is equivalent to Problem \ref{p:frac_opt}.

{
As a summary, Problem \ref{p:raw_opt_orig} is equivalent with Problem \ref{p:raw_opt} according to Lemma \ref{lem:indicator_remove}, and Problem \ref{p:raw_opt} is equivalent with Problem\ref{p:frac_opt} by Lemma \ref{lem:S2beta}.
Also with the equivalence between Problem \ref{p:frac_opt} and Problem \ref{p:upper_level_raw} as just disclosed, we can claim the equivalence between Problem \ref{p:raw_opt_orig} and Problem \ref{p:upper_level_raw}.
It should be also noticed that the role of Problem \ref{p:lower_level_raw} is to return the value of Problem \ref{p:upper_level_raw}'s cost function for an input of $a$.
}

\section{Optimal Solution for the Lower-Level Problem (Problem \ref{p:lower_level_raw})} \label{s:optimal_solution_beta_b}
For the lower-level optimization problem, i.e., Problem \ref{p:lower_level_raw}, its cost function is a joint convex function of $\{b_k\}$ and $\{\beta_k\}$, {and its feasible solution is also convex.}
Hence Problem \ref{p:lower_level_raw} is a convex optimization problem. In addition, it also satisfies Slater's condition. In this case, Karush-Kuhn-Tucker (KKT) condition can serve as a necessary and sufficient condition of the optimal solution of Problem \ref{p:lower_level_raw} {\cite{An2022Joint}}. Specifically, the KKT condition can be given as follows
\begin{subequations} \label{e:KKT}
\begin{align}
-2 (a b_k h_k - \beta_k) {c_k} + \lambda - z_k + y_k =0, \forall k \in \mathcal{K}, \label{e:KKT_diff_beta}\\
2 (a b_k h_k - \beta_k) {c_k} a h_k + \mu_k - \gamma_k = 0, \forall k \in \mathcal{K}, \label{e:KKT_diff_b}\\
0 \leq \beta_k \leq \beta_k^{\max}, \forall k \in \mathcal{K}, \label{e:KKT_box_beta} \\
0 \leq b_k \leq b_k^{\max}, \forall k \in \mathcal{K},  \label{e:KKT_box_b}\\
\sum_{k=1}^K \beta_k = 1, \label{e:KKT_sum_beta} \\
z_k \beta_k = 0, \forall k \in \mathcal{K}, \label{e:KKT_z_k}\\
y_k (\beta_k - \beta_k^{\max}) = 0, \forall k \in \mathcal{K}, \label{e:KKT_y_k}\\
\gamma_k b_k = 0, \forall k \in \mathcal{K}, \label{e:KKT_gamma_k}\\
\mu_k (b_k - b_k^{\max}) = 0, \forall k \in \mathcal{K}, \label{e:KKT_mu_k}\\
z_k \geq 0; y_k \geq 0; \gamma_k \geq 0; \mu_k \geq 0, \forall k \in \mathcal{K} \label{e:KKT_Lagrange_nonnegative}
\end{align}
\end{subequations}
where $z_k$, $y_k$, $\gamma_k$, and $\mu_k$ are non-negative Lagrange multipliers associated with the constraints $\beta_k\geq 0$, $\beta_k \leq \beta_{\max}$, $b_k \geq 0$, and $b_k \leq b_k^{\max}$, respectively, and $\lambda$ is the Lagrange multiplier associated with the constraint in (\ref{e:KKT_sum_beta}).

\subsection{Characterizing Candidate Optimal Solutions}
We will first find optimal solutions of Problem \ref{p:lower_level_raw}. To achieve this, special properties of an optimal solution of Problem \ref{p:lower_level_raw} will be analyzed based on listed KKT condition in (\ref{e:KKT}).
From (\ref{e:KKT_diff_beta}) in the KKT condition, we have
\begin{equation} \label{e:KKT_diff_beta_2}
\left(\lambda - z_k + y_k\right)= 2 {c_k} \left(a h_k b_k - \beta_k \right), \forall k \in \mathcal{K}.
\end{equation}
Substituting the expression of $2 \left(a h_k b_k - \beta_k \right)$ in (\ref{e:KKT_diff_beta_2}) into (\ref{e:KKT_diff_b}), we have
\begin{equation} \label{e:a_h_k_eqn}
\left(\lambda - z_k + y_k\right) \times (a h_k)  = \left(\gamma_k - \mu_k\right), \forall k \in \mathcal{K}.
\end{equation}

Next, we investigate two cases: 1) $\lambda=0$; 2) $\lambda \neq 0$.
\subsubsection{Case with $\lambda=0$}
When $\lambda =0$, equation (\ref{e:a_h_k_eqn}) becomes
\begin{equation} \label{e:a_h_k_eqn_lambda0}
\left(- z_k + y_k\right) \times (a h_k)  = \left(\gamma_k - \mu_k\right), \forall k \in \mathcal{K},
\end{equation}
which implies that the term $\left(- z_k + y_k\right) $ and the term $\left(\gamma_k - \mu_k\right)$ should be both positive or both negative. In addition, from (\ref{e:KKT_z_k}) and (\ref{e:KKT_y_k}), it can be seen that at least one of $z_k$ and $y_k$ (which are nonnegative) should be zero.
Similarly, from (\ref{e:KKT_gamma_k}) and (\ref{e:KKT_mu_k}), at least one of $\gamma_k$ and $\mu_k$ (which are nonnegative) should be zero.
We have three possible scenarios as follows.
\begin{itemize}
\item Scenario I: $\left(\gamma_k - \mu_k\right) > 0$. Together with the fact that at least one of $\gamma_k$ and $\mu_k$ (which are nonnegative) should be zero, we should have $\gamma_k>0$ and $\mu_k = 0$. Then from (\ref{e:KKT_gamma_k}), we have $b_k=0$.
{On the other hand, $\left(\gamma_k - \mu_k\right) > 0$ also implies that $\left(-z_k + y_k \right) >0$ according to (\ref{e:a_h_k_eqn_lambda0}), which further implies that $y_k>0$ and $z_k=0$. In this case, there is $\beta_k = \beta_k^{\max}$ according to (\ref{e:KKT_y_k}).
So in this scenario, there is $b_k=0$ and $\beta_k = \beta_k^{\max}$.
However, the MSE in this scenario can be further reduced by setting $b_k = \min \left(b_k^{\max}, \frac{\beta_k^{\max}}{a h_k} \right) >0$ with $\beta_k = \beta_k^{\max}$.
Therefore, the optimal solution does not happen in this scenario and we discard it.
}
\item Scenario II: $\left(\gamma_k - \mu_k\right) < 0$.
{Similar to the discussion in Scenario I, we have $\gamma_k=0$, $\mu_k>0$, $y_k=0$, and $z_k>0$. Correspondingly, there is $b_k = b_k^{\max}$ according to (\ref{e:KKT_mu_k}) and $\beta_k = 0$ according to (\ref{e:KKT_z_k}). However, the MSE in this case can be further reduced by setting $b_k$ as 0 rather than $b_k^{\max}$ when $\beta_k = 0$. Hence this scenario is also discarded.}
\item Scenario III: $\left(\gamma_k - \mu_k\right) = 0$. Since at least one of $\gamma_k$ and $\mu_k$ (which are nonnegative) should be zero, we have $\gamma_k = \mu_k=0$. {Similar to the discussion in Scenario I and II}, we also have $z_k = y_k = 0$. From (\ref{e:KKT_diff_beta_2}), we have
\begin{equation} \label{e:beta_ahb_eqn_lambda_0}
\beta_k = a h_k b_k, \forall k \in \mathcal{K},
\end{equation}
referred to as {\bf Candidate Solution I} for Problem \ref{p:lower_level_raw}.
It can be seen that when $\beta_k$ and $b_k$ satisfies (\ref{e:beta_ahb_eqn_lambda_0}), the KKT condition is satisfied, and the cost function of Problem \ref{p:lower_level_raw} achieves its minimal value $a^2 \sigma^2$.
\end{itemize}

\subsubsection{Case with $\lambda \neq 0$} We investigate three scenarios: $(\lambda - z_k + y_k)=0$, $(\lambda - z_k + y_k)>0$, and $(\lambda - z_k + y_k)<0$.

\begin{itemize}
\item Scenario with $(\lambda - z_k + y_k)=0$.
{In this scenario, similar with the discussion in Scenario III for the case with $\lambda=0$, there is $\gamma_k= \mu_k =0$. Since $(\lambda - z_k + y_k)=0$,  there is $z_k = \left(\lambda + y_k\right)$. As a non-negative Lagrange multiplier,
$z_k$ could be $z_k =0$ and $z_k >0$. }
\begin{itemize}
\item {If $z_k =0$}, we have $\left(\lambda + y_k \right)=0$. Since $\lambda \neq 0$, we have $y_k \neq 0$. As $y_k$ is a non-negative Lagrange multiplier, we have $y_k >0$, which further implies that $\beta_k = \beta_k^{\max}$ by (\ref{e:KKT_y_k}). From (\ref{e:KKT_diff_beta_2}), we have $ \left(a h_k b_k - \beta_k \right) = \frac{1}{2 {c_k} }\left(\lambda - z_k + y_k\right) = 0$. Hence $b_k = \frac{\beta_k}{a h_k} = \frac{\beta_k^{\max}}{a h_k}$. This is called {\bf Candidate Solution II} for Problem \ref{p:lower_level_raw}. Note that for Candidate Optimal Solution II, we have $\lambda = \left(z_k - y_k\right) < 0$.
Moreover, the holding of $b_k = \frac{\beta_k^{\max}}{a h_k}$ implies that $\frac{\beta_k^{\max}}{a h_k} \leq b_k^{\max}$, i.e., $a h_k b_k^{\max} \geq \beta_k^{\max}$.
\item

{If $z_k > 0$, we have $\beta_k =0$ according to (\ref{e:KKT_z_k}), which further implies $b_k=0$ for minimizing the term $(a b_k h_k - \beta_k)^2 c_k$ in the expression of MSE. This is called {\bf Candidate Solution III} for Problem \ref{p:lower_level_raw}. With $b_k= 0$, there is $y_k=0$ according to (\ref{e:KKT_y_k}). Given that $\left(\lambda - z_k + y_k \right) a h_k = \left(\gamma_k - \mu_k\right) = 0$ in this case, and recall the fact that $z_k > 0$ and $y_k =0$, there is $\lambda >0$.}
\end{itemize}

\item Scenario with $(\lambda - z_k + y_k)<0$, which indicates $\left(\gamma_k -\mu_k\right)<0$ according to (\ref{e:a_h_k_eqn}). Recalling the at least one of $\gamma_k$ and $\mu_k$ (which are nonnegative) should be zero, we have $\gamma_k =0$ and $\mu_k > 0$, which further indicates that $b_k = b_k^{\max}$ from (\ref{e:KKT_mu_k}). {In this scenario, $z_k$ could be $z_k >0$ or $z_k =0$.}
\begin{itemize}
\item
{
If $z_k=0$, we discuss situations of $y_k=0$ and $y_k>0$, respectively.}
\begin{itemize}
\item $y_k > 0$: In this situation, we have $\beta_k = \beta_k^{\max}$ according to (\ref{e:KKT_y_k}). This is called {{\bf Candidate Solution IV}} for Problem \ref{p:lower_level_raw}. With this candidate optimal solution, according to (\ref{e:KKT_diff_beta_2}), $ \left( a h_k b_k^{\max} - \beta_k^{\max} \right) = \frac{(\lambda - z_k + y_k)}{2 {c_k}} < 0$. Hence there is a requirement that $a h_k < \frac{\beta_k^{\max}}{b_k^{\max}}$, i.e., $a h_k b_k^{\max} < \beta_k^{\max}$. Moreover, the inequality $\frac{(\lambda - z_k + y_k)}{2 {c_k}} < 0$, together with the fact that $z_k=0$ and $y_k>0$, indicate $\lambda<0$.
\item $y_k =0$: In this situation, $\beta_k$ can range from 0 to $\beta_k^{\max}$. This is called {{\bf Candidate Solution V}} for Problem \ref{p:lower_level_raw}. With this candidate optimal solution, from the assumption $(\lambda - z_k + y_k)<0$, it can be easily seen that $\lambda <0$. Also from (\ref{e:KKT_diff_beta_2}), there is $ \left( a h_k b_k^{\max} - \beta_k \right) = \frac{(\lambda - z_k + y_k)}{2 {c_k}} < 0$, which implies that $\beta_k > a h_k b_k^{\max}$. This also means the inequality $a h_k b_k^{\max} < \beta_k^{\max}$ holds.
\end{itemize}
\item
{
If $z_k >0$, from (\ref{e:KKT_z_k}) we have $\beta_k = 0$, which further implies $b_k=0$ for minimizing the term $(a b_k h_k - \beta_k)^2 c_k$ in the expression of MSE. On the other hand, $\mu_k$ has to be zero according to (\ref{e:KKT_mu_k}) for $b_k =0$, which contradicts with the fact that $\mu_k>0$ claimed at the beginning of this scenario.
}
\end{itemize}

\item Scenario with $(\lambda - z_k + y_k)>0$. Similar to the scenario with $(\lambda - z_k + y_k)<0$, it can be derived that $\gamma_k > 0$ and $\mu_k =0$. Together with (\ref{e:KKT_gamma_k}), we have $b_k = 0$.

{Then we have $2c_k \left(a h_k b_k - \beta_k \right) = (\lambda - z_k + y_k) >0$ according to (\ref{e:KKT_diff_beta_2}), which contradicts with the fact that $\beta_k \geq 0$. So this scenario is also discarded.
}
\end{itemize}

\begin{remark} \label{r:possible_solution}
{Overall, there are 5 candidate solutions, which are summarized as follows}
\begin{itemize}
\item Candidate Solution I:  $\lambda =0$, $\gamma_k = \mu_k = z_k = y_k = 0$. $b_k \in [0, b_k^{\max}]$, $\beta_k \in [0, \beta_k^{\max}]$, and $a h_k b_k = \beta_k $ for $k\in \mathcal{K}$.
\item Candidate Solution II: $\lambda < 0$, $\gamma_k = \mu_k = z_k=0$, $y_k > 0$, and $\left(\lambda - z_k + y_k\right) = 0$. $\beta_k = \beta_k^{\max}$ and $b_k = \frac{\beta_k^{\max}}{a h_k}$. This solution happens when $a h_k b_k^{\max} \geq \beta_k^{\max}$.
\item {Candidate Solution III: $\lambda > 0$, $\gamma_k = \mu_k = y_k=0$, $z_k > 0$, and $\left(\lambda - z_k + y_k\right) = 0$. $\beta_k = 0$ and $b_k = 0$. }
\item {Candidate Solution IV}: $\lambda < 0$, $\gamma_k  = z_k=0$, $\mu_k > 0$, $y_k > 0$, and $\left(\lambda - z_k + y_k\right) < 0$. $b_k = b_k^{\max}$ and $\beta_k = \beta_k^{\max}$. This solution happens when $a h_k b_k^{\max} < \beta_k^{\max}$.
\item {Candidate Solution V}: $\lambda< 0$, $\gamma_k = z_k = y_k =0$, $\mu_k >0$, and $\left(\lambda - z_k + y_k\right) < 0$. $b_k = b_k^{\max}$ and $\beta_k \in \left(a h_k b_k^{\max},  \beta_k^{\max}\right]$. This solution happens when $a h_k b_k^{\max} < \beta_k^{\max}$.
\end{itemize}
\end{remark}

It can be seen that, if $\lambda = 0$, then all mobile devices take Candidate Solution I, in which we still need to determine the exact values of $b_k$ and $\beta_k$ for the $k$th mobile device.
If $\lambda<0$, then the $k$th mobile device takes Candidate Solution II if $a h_k b_k^{\max} \geq \beta_k^{\max}$, and takes Candidate Solution IV or V if $a h_k b_k^{\max} < \beta_k^{\max}$.
{If $\lambda>0$, only Candidate Solution III is active and all the other candidate solutions (including Candidate Solutions I, II, IV, and V) are precluded because the associated $\lambda$ values of these candidate solutions are either zero or positive.
On other hand, Candidate Solution III cannot work for every $k\in \mathcal{K}$ since $\sum_{k=1}^{K} \beta_k$ should be 1 for Problem \ref{p:lower_level_raw} while $\beta_k = 0$ in Candidate Solution III.
So we can omit Candidate Solution III.}

{For the other four candidate solutions with $\lambda = 0$ or $\lambda>0$, i.e., Candidate Solutions I, II, IV, and V, since we only need {\it one} optimal solution of Problem \ref{p:lower_level_raw},} we will first try $\lambda<0$. If each mobile device's associated solution (i.e., Candidate Solution II, IV, or V) is feasible, then the mobile devices' solutions form optimal solution of Problem \ref{p:lower_level_raw}; otherwise, we try $\lambda=0$ with Candidate Solution I.

\subsection{Finding An Optimal Solution of Problem \ref{p:lower_level_raw}}\label{S:lower_level_raw}

We partition set $\mathcal{K}$ into two disjoint subsets, $\mathcal{K}_1$ and $\mathcal{K}_2$ such that $\mathcal{K} = \mathcal{K}_1 \cup \mathcal{K}_2$, $\mathcal{K}_1 = \{k | a h_k b_k^{\max} \geq \beta_k^{\max}, k \in \mathcal{K}\}$, and $\mathcal{K}_2 = \{k | a h_k b_k^{\max} < \beta_k^{\max}, k \in \mathcal{K}\}$. Consider $\lambda<0$. As aforementioned, for $k \in \mathcal{K}_1$, the $k$th mobile device's solution is $\beta_k = \beta_k^{\max}$ and $b_k = \frac{\beta_k^{\max}}{a h_k}$ (Candidate Solution II); for $k \in \mathcal{K}_2$, the $k$th mobile device's solution is $b_k = b_k^{\max}$, $\beta_k = \beta_k^{\max}$ (Candidate Solution IV) or $\beta_k \in \left(a h_k b_k^{\max},  \beta_k^{\max}\right]$ (Candidate Solution V). We need to verify whether or not a $\lambda$ exists to make all these happen.

From (\ref{e:KKT_diff_beta_2}) and the facts that $b_k = b_k^{\max}$ and $z_k=0$ for $k \in \mathcal{K}_2$, $\beta_k$ can be written as a function with $\lambda$, which is given as
\begin{equation} \label{e:beta_lambda_express_raw}
\beta_k(\lambda) =  \left( a h_k b_k^{\max} - \frac{\lambda }{2 {c_k}} - \frac{y_k}{2 {c_k}} \right), \forall k \in \mathcal{K}_2.
\end{equation}
According to (\ref{e:beta_lambda_express_raw}) and the discussion for Candidate Solution IV and Candidate Solution V, for $k\in \mathcal{K}_2$, when $\beta_k \in \left(a h_k b_k^{\max},  \beta_k^{\max}\right]$, we have $y_k=0$ and $\beta_k = \left( a h_k b_k^{\max} - \frac{\lambda }{2{c_k}} \right) = \min\left(\left( a h_k b_k^{\max} - \frac{\lambda }{2{c_k}} \right), \beta_k^{\max}\right)$;
when $\beta_k = \beta_k^{\max}$, we have $y_k \geq 0$ and $\beta_k = \left( a h_k b_k^{\max} - \frac{\lambda }{2{c_k}} - \frac{y}{2{c_k}} \right) \leq \left( a h_k b_k^{\max} - \frac{\lambda }{2{c_k}} \right)$, and thus, $\beta_k$ can also be written as $\beta_k=\min\left(\left( a h_k b_k^{\max} - \frac{\lambda }{2 {c_k}} \right), \beta_k^{\max}\right)$.  In summary,
the expression of $\beta_k$ with $\lambda$ can be simplified to be
\begin{equation} \label{e:beta_lambda_express}
\beta_k(\lambda) = \min \left(\left( a h_k b_k^{\max} - \frac{\lambda }{2 {c_k}} \right), \beta_k^{\max}\right), \forall k \in \mathcal{K}_2,
\end{equation}
which is a non-increasing function with $\lambda$.

Substituting (\ref{e:beta_lambda_express}) in equality (\ref{e:sum_beta}), we can get
\begin{equation} \label{e:beta_lambda_sum_1}
 \sum_{k \in \mathcal{K}_1} \beta_k^{\max} + \sum_{k \in \mathcal{K}_2} \beta_k(\lambda) = 1.
\end{equation}
For equality (\ref{e:beta_lambda_sum_1}), it can be found that every term on its left-hand side is a non-increasing function with $\lambda$. Hence the solution, denoted $\lambda^*$, for the equation in (\ref{e:beta_lambda_sum_1}) can be found through a bisection search method for $\lambda  \in \left(\lambda_{\min} ,0\right)$,
where
\begin{equation}
\lambda_{\min} \triangleq  \mathop{\min} \limits_{k\in \mathcal{K}_2}  \quad 2 {c_k} \left(a h_k b_k^{\max} - \beta_k^{\max} \right)
\end{equation}
Then, the optimal $\beta_k$ for $k\in \mathcal{K}_2$ can be calculated according to (\ref{e:beta_lambda_express}).

The above solution of $b_k$ and $\beta_k$ for the $k$th mobile device (also solution of Problem \ref{p:lower_level_raw}) is under the condition that (\ref{e:beta_lambda_sum_1}) has a solution of $\lambda$. However, when (\ref{e:beta_lambda_sum_1}) does not have a solution of $\lambda$, we need to find optimal solution of Problem \ref{p:lower_level_raw} from Candidate Solution I, as follows.

Equality (\ref{e:beta_lambda_sum_1}) does not have a solution of $\lambda$ when one of the following two events happens:
1) $\sum_{k\in \mathcal{K}_1} \beta_k^{\max} > 1$;
2) $\sum_{k\in \mathcal{K}_1} \beta_k^{\max} <1$ but there does not exist a $\lambda \in \left(\lambda_{\min}, 0\right)$ to make the equality (\ref{e:beta_lambda_sum_1}) hold.

First consider the case with $\sum_{k\in \mathcal{K}_1} \beta_k^{\max} > 1$.
Denote {$\delta = \min\left(\Delta,  \mathop{\min} \limits_{k\in \mathcal{K}_2}  a h_k b_k^{\max} \right)$, where $\Delta$ is any positive value less than 1}. From definition of $\mathcal{K}_2$, we have $0<\delta<1$. Since $\sum_{k\in \mathcal{K}_1} \beta_k^{\max} > 1$, we can select a set of $\beta_k\in (0, \beta_k^{\max}]$ for $k \in \mathcal{K}_1$ such that $\sum_{k \in \mathcal{K}_1} \beta_k = 1-\delta$, and select $b_k=\frac{\beta_k}{ah_k} \in (0, b_k^{\max}]$ for $k\in \mathcal{K}_1$. It can be seen that $\sum_{k \in \mathcal{K}_1} (ah_kb_k-\beta_k)^2=0$. We then select a set of $\beta_k\in (0, \beta_k^{\max})$ for $k \in \mathcal{K}_2$ such that $\sum_{k \in \mathcal{K}_2} \beta_k = \delta$, and select $b_k=\frac{\beta_k}{ah_k} \in (0, b_k^{\max}]$ for $k\in \mathcal{K}_2$. It can be seen that $\sum_{k \in \mathcal{K}_2} (ah_kb_k-\beta_k)^2=0$. By the above setting (which is actually Candidate Solution I as (\ref{e:beta_ahb_eqn_lambda_0}) is held), the cost function of Problem \ref{p:lower_level_raw} achieves its optimal value $a^2 \sigma^2$ since $\sum_{k \in \mathcal{K}} (ah_kb_k-\beta_k)^2=0$ in this setting
\footnote{{It should be noticed that the definitions of $\Delta$ and $\delta$ are necessary for this case, i.e., $\sum_{k\in \mathcal{K}_1} \beta_k^{\max} >1$.
According to the definition of $\delta$, it is no larger than both $\Delta$ and $\mathop{\min} \limits_{k\in \mathcal{K}_2} a h_k b_k^{\max}$.
The role of $\Delta$ is to bound $\delta $ by 1, which guarantees that both $\sum_{k \in \mathcal{K}_2} \beta_k = \delta$ and $\sum_{k \in \mathcal{K}_1} \beta_k = (1 - \delta)$ to be no larger than 1.
The term $\mathop{\min} \limits_{k\in \mathcal{K}_2} a h_k b_k^{\max}$ in the definition of $\delta$ is to make sure the derived $b_k=\beta_k/a h_k$ to be no larger than $b_k^{\max}$ for $k\in \mathcal{K}_2$, considering that $\beta_k \leq \sum_{k \in \mathcal{K}_2} \beta_k = \delta \leq \mathop{\min} \limits_{k\in \mathcal{K}_2} a h_k b_k^{\max} \leq a h_k b_k^{\max}$ for $k\in \mathcal{K}_2$.
The selection of $\Delta$ and $\delta$ could be not unique, which implies the optimal solution of Problem \ref{p:lower_level_raw} in this case could be not unique either.
This is normal for Problem \ref{p:lower_level_raw} to achieve its minimal cost function $a^2 \sigma^2$, which only requires $ah_k b_k$ to fully offset $\beta_k$ for every $k\in \mathcal{K}$.
With the above selection of $\delta$ and $\Delta$, it is possible to find multiple sets of $b_k$ and $\beta_k$ to make the offset between $ah_k b_k$ and $\beta_k$ happen for every $k\in \mathcal{K}$ while fulfilling all the constraints of Problem \ref{p:lower_level_raw}.}}
.

Now we investigate the case $\sum_{k\in \mathcal{K}_1} \beta_k^{\max} <1$ but there does not exist a $\lambda \in \left(\lambda_{\min}, 0\right)$ to make the equality (\ref{e:beta_lambda_sum_1}) hold. This will happen when the left-hand function of (\ref{e:beta_lambda_sum_1}) is no less than 1 even for $\lambda= 0$.
For such a case, we first need to figure out when it will happen. By setting $\lambda$ to be zero,
(\ref{e:beta_lambda_sum_1})  turns to be
\begin{equation}
 \sum_{k \in \mathcal{K}_1} \beta_k^{\max} + \sum_{k \in \mathcal{K}_2} a h_k b_k^{\max} = 1,
\end{equation}
whose solution of $a$, denoted as $a_{\text{th}}$, is given as
\begin{equation} \label{e:a_th_exp}
a_{\text{th}} \triangleq  \frac{\left(1- \sum_{k \in \mathcal{K}_1} \beta_k^{\max} \right)}{\sum_{k \in \mathcal{K}_2} h_k b_k^{\max}}.
\end{equation}
It can be found that when $a \geq a_{\text{th}}$, we cannot find a solution of $\lambda \in (\lambda_{\min}, 0)$ to satisfy the equality (\ref{e:beta_lambda_sum_1}). In other words, when $a \geq a_{\text{th}}$, Candidate Solution II, IV, and V cannot serve as an optimal solution of Problem \ref{p:lower_level_raw}, and we need to resort to Candidate Solution I, which implies the holding of (\ref{e:beta_ahb_eqn_lambda_0}). The procedure to work out the optimal $b_k$ and $\beta_k$ of Problem \ref{p:lower_level_raw} for $k\in \mathcal{K}$ can be given as follows.
For $k\in \mathcal{K}_1$, set $\beta_k = \beta_k^{\max}$, and $b_k = \frac{\beta_k^{\max}}{a h_k}$.
For $k\in \mathcal{K}_2$, set $b_k = \frac{a_{\text{th}}  b_k^{\max}}{a}$ (which is applicable since $\frac{a_{\text{th}}}{a} \leq 1$ for $a \geq a_{\text{th}}$) and $\beta_k= a h_k b_k = a_{\text{th}} h_k  b_k^{\max}$ (which is also applicable since $a h_k b_k \leq a h_k b_k^{\max} < \beta_{k}^{\max}$ for $k\in \mathcal{K}_2$).
Recalling the expression of $a_{\text{th}}$ in (\ref{e:a_th_exp}), it can be checked that
\begin{equation}
\begin{array}{ll}
\sum_{k\in \mathcal{K}} \beta_k  & = \sum_{k\in \mathcal{K}_1} \beta_{k}^{\max} + \sum_{k\in \mathcal{K}_2} a_{\text{th}} h_k  b_k^{\max}  \\
& = \sum_{k\in \mathcal{K}_1} \beta_{k}^{\max} + \left( 1- \sum_{k\in \mathcal{K}_1} \beta_{k}^{\max}\right) \\
& =1.
\end{array}
\end{equation}
With the above optimal solution of $b_k$ and $\beta_k$ for $k\in \mathcal{K}$ when $a \geq a_{\text{th}}$ , the cost function of Problem \ref{p:lower_level_raw} achieves its optimal value $a^2 \sigma^2$.

Overall, the method for finding an optimal solution of Problem \ref{p:lower_level_raw} is given in Algorithm \ref{a:lower_level}.

\begin{algorithm}
	\caption{The procedure of finding an optimal solution of Problem \ref{p:lower_level_raw}.} \label{a:lower_level}
	\KwIn{The variable $a$. The variables $h_k$, $\beta_k^{\max}$, and $b_k^{\max}$ for $k\in \mathcal{K}$.}
  \KwOut{An optimal solution of $\beta_k$ and $b_k$, $\forall k \in \mathcal{K}$, for Problem \ref{p:lower_level_raw}.}
  {Define the set $\mathcal{K}_1$ and $\mathcal{K}_2$, such that $\mathcal{K}_1 = \{k | a h_k b_k^{\max} \geq \beta_k^{\max}, k \in \mathcal{K}\}$ and $\mathcal{K}_2 = \{k | a h_k b_k^{\max} < \beta_k^{\max}, k \in \mathcal{K}\}$.
  Calculate $a_{\text{th}}$ according to (\ref{e:a_th_exp}).}

  \eIf{$\sum_{k\in \mathcal{K}_1} \beta_k^{\max} \leq 1$}{
    \eIf{$a < a_{\text{th}}$}{
      Use bisection-search method to find the $\lambda \in [\lambda_{\min},0)$ satisfying the equality in (\ref{e:beta_lambda_sum_1}), denoted as $\lambda^*$.

      For $k \in \mathcal{K}_1$, set $\beta_k = \beta_k^{\max}$ and $b_k = \frac{\beta_k^{\max}}{a h_k}$.

      For $k \in \mathcal{K}_2$, set $b_k = b_k^{\max}$ and $\beta_k = \beta_k(\lambda^*)$.
    }
    {
      For $k \in \mathcal{K}_1$, set $\beta_k = \beta_k^{\max}$ and $b_k = \frac{\beta_k^{\max}}{a h_k}$.

      For $k \in \mathcal{K}_2$, set $b_k = \frac{a_{\text{th}}  b_k^{\max}}{a}$ and $\beta_k = a_{\text{th}} h_k  b_k^{\max}$.
    }
  }
  {
    Select a set of $\beta_k\in (0, \beta_k^{\max}]$ for $k \in \mathcal{K}_1$ such that $\sum_{k \in \mathcal{K}_1} \beta_k = 1-\delta$, and select $b_k=\frac{\beta_k}{ah_k} \in (0, b_k^{\max}]$ for $k\in \mathcal{K}_1$. Here {$\delta = \min\left(\Delta,  \mathop{\min} \limits_{k\in \mathcal{K}_2}  a h_k b_k^{\max} \right)$ and $\Delta \in (0,1)$}.

    Select a set of $\beta_k\in (0, \beta_k^{\max})$ for $k \in \mathcal{K}_2$ such that $\sum_{k \in \mathcal{K}_2} \beta_k = \delta$, and select $b_k=\frac{\beta_k}{ah_k} \in (0, b_k^{\max}]$ for $k\in \mathcal{K}_2$.
  }

\end{algorithm}

For computation complexity of Algorithm \ref{a:lower_level}, the worst case happens when $\sum_{k\in \mathcal{K}_1} \beta_k^{\max} \leq 1$ and $a < a_{\text{th}}$. The computation burden comes from the bisection search of $\lambda^*$ and the calculation of $\beta_k$ and $b_k$ for $k\in \mathcal{K}$, whose complexity can be written as $O\left(\log\left(\frac{\lambda_{\min}}{\delta_B}\right) K\right)$, where $\delta_B$ is the tolerance for bisection search.

\section{Optimal Solution for Upper-Level Problem (Problem \ref{p:upper_level_raw})} \label{s:optimal_solution_a}

In this section, the optimal $a$ of Problem \ref{p:upper_level_raw}, which is also optimal for Problem \ref{p:raw_opt}, is to be found.
{However, the convexity of $E(a)$, which is the cost function of Problem \ref{p:upper_level_raw}, is unclear and hard to explore. This challenge will be overcome by our proposed solution given in the following.}

As the value of $a$ grows, it can be checked that the set $\mathcal{K}_1$ enlarges and the set $\mathcal{K}_2$ shrinks. Hence both set $\mathcal{K}_1$ and set $\mathcal{K}_2$ are functions of $a$, which are denoted as $\mathcal{K}_1(a)$ and $\mathcal{K}_2(a)$, respectively, in this section.
Similarly the $\lambda^*$ solving the equation in (\ref{e:beta_lambda_sum_1}) is also a function of $a$, which is denoted as $\lambda^*(a)$ in this section.

As $a$ grows, the cardinality of $\mathcal{K}_1(a)$ will enlarge from 0 to $K$, while the cardinality of $\mathcal{K}_2(a)$ will shrink from $K$ to 0 at the same time.
Then there would be multiple intervals of $a$. When $a$ varies within any interval, the cardinality of the set $\mathcal{K}_1(a)$ and $\mathcal{K}_2(a)$ keep unchanged. When $a$ grows from one interval to the next interval, the cardinality of the set $\mathcal{K}_1(a)$ and $\mathcal{K}_2(a)$ will change.
To characterizing these intervals, the boundaries of every interval of $a$ need to be found.

Define $s_k \triangleq \frac{\beta_k^{\max}}{h_k b_k^{\max}}$ for $k\in \mathcal{K}$ and sort $s_k$ for $k\in \mathcal{K}$ in ascending order, such that $s_{(1)} \leq s_{(2)} \leq s_{(3)} \cdots \leq s_{(K)}$, where $s_{(k)} \in \{s_1, s_2, \cdots, s_K\}$ for $k \in \mathcal{K}$.
\begin{itemize}
\item When $ a \in (0, s_{(1)})$, $|\mathcal{K}_1(a)| = 0$ and $|\mathcal{K}_2(a)|  = K$.
\item When $ a \in [s_{(k)}, s_{(k+1)})$, $|\mathcal{K}_1(a)|= k$ and $|\mathcal{K}_2(a)|  = \left(K-k\right)$ for $k \in \mathcal{K} \setminus \{K\}$.
\item When $ a \in [s_{(K)}, +\infty)$, $|\mathcal{K}_1(a)|= K$ and $|\mathcal{K}_2(a)|  = 0$.
\end{itemize}
There are $\left(K+1\right)$ intervals of $a$, given as $(0, s_{(1)}), [s_{(1)}, s_{(2)}), \cdots, [s_{(K-1)}, s_{(K)}), [s_{(K)}, \infty)$. Within these $\left(K+1\right)$ intervals, $|\mathcal{K}_1(a)|$ is from 0 to $K$, while $|\mathcal{K}_2(a)|$ is from $K$ to 0.

Define $a^{\max}$ as the minimal $a$ such that $\sum_{k \in \mathcal{K}_1(a)}  \beta_k^{\max}  > 1$.
When $a \geq a^{\max}$, according to the discussion in the preceding section, the minimal cost of Problem \ref{p:lower_level_raw}, which is also the cost function of Problem \ref{p:upper_level_raw}, is always $a ^2 \sigma^2$. In this case, to minimize the cost function of Problem \ref{p:upper_level_raw}, it is optimal to set $a$ as $a = a^{\max}$.
It can be checked that $a^{\max} \leq s_{(K)}$ since $\sum_{k \in \mathcal{K}_1(s_{(K)})} \beta_k^{\max} = \sum_{k \in \mathcal{K}} \beta_k^{\max} \geq 1$.

Collecting the above discussions about $a$'s intervals and supposing $a^{\max} \in [s_{(M)}, s_{(M+1)}]$ where $M \leq  \left(K -1\right)$, the intervals of $a$ to be investigated are $(0, s_{(1)})$, $[s_{(1)}, s_{(2)})$, $\cdots$, $[s_{(M)}, a^{\max})$, $[a^{\max}, +\infty)$, whose total number may reach up to $(K+1)$ at most.
To facilitate the discussion in the following, these intervals are indexed as $\mathcal{A}_0$, $\mathcal{A}_1, \cdots, \mathcal{A}_{M+1}$, respectively.

When $a$ is in one interval $\mathcal{A}_m$ ($m =0, 1, ..., M$), the $a_{\text{th}}$ defined in (\ref{e:a_th_exp}) is a fixed number and can be denoted as $a_{\text{th},m}$.
Recalling that for $a \geq  a_{\text{th},m}$, there is no feasible solution of $\lambda^*(a) $ for solving equality (\ref{e:beta_lambda_sum_1}), in which case the minimal achievable cost function of Problem \ref{p:lower_level_raw} is $a^2 \sigma^2$. Then we can divide the interval $\mathcal{A}_m$ into two intervals, $\mathcal{B}_m$ and $\mathcal{C}_m$.
For the ease of discussion in the following, suppose the left-end point and the right-end point of interval $\mathcal{A}_{m}$ are $l_{m}^a$ and $r_{m}^a$, respectively.
Then $\mathcal{B}_m \triangleq  [l_{m}^a, a_{\text{th},m})$ and $\mathcal{C}_m \triangleq [a_{\text{th},m}, r_m^a)$, which indicates that $\mathcal{A}_m = \mathcal{B}_m \cup \mathcal{C}_m$ and $\mathcal{B}_m \cap \mathcal{C}_m = \emptyset $.

To find the global optimal $a$ for Problem \ref{p:upper_level_raw}, we need to find the optimal $a$ within every interval $\mathcal{A}_m$ for $m=0, 1, ..., M+1$, and select the best one among all the intervals. Recall that the optimal $a$ in $\mathcal{A}_{M+1}$ is $a^{\max}$. So next we focus on intervals $\mathcal{A}_0, \mathcal{A}_1,...,\mathcal{A}_M$.

Within one interval $\mathcal{A}_m = \mathcal{B}_m \cup \mathcal{C}_m$ ($m=0,1,...,M$), when $a\in \mathcal{C}_m$, it can be found that the cost function of Problem \ref{p:upper_level_raw} is $a^2 \sigma^2$ and the minimal achievable cost function of Problem \ref{p:upper_level_raw} is $a_{\text{th}, m}^2 \sigma^2$; when $a \in \mathcal{B}_m$, we need to characterize the optimal solution of $a$  as follows.

When $a$ is within one interval $\mathcal{B}_m$, the set $\mathcal{K}_1(a)$ and $\mathcal{K}_2(a)$ keep unchanged.
We first investigate the set $\mathcal{K}_2(a)$.
For the set $\mathcal{K}_2(a)$, it can be further decomposed into two sets, $\mathcal{K}_2^{\text{I}}(a)$ and $\mathcal{K}_2^{\text{II}}(a)$, such that $\mathcal{K}_2^{\text{I}}(a) = \left\{ k | b_k = b_k^{\max}, \beta_k = \beta_k^{\max}, k \in \mathcal{K}_2(a)\right\}$ and $\mathcal{K}_2^{\text{II}}(a) = \left\{ k | b_k = b_k^{\max}, \beta_k < \beta_k^{\max}, k \in \mathcal{K}_2(a)\right\}$.
It can be found that $\mathcal{K}_2(a) = \mathcal{K}_2^{\text{I}}(a) \cup \mathcal{K}_2^{\text{II}}(a)$ and $\mathcal{K}_2^{\text{I}}(a) \cap \mathcal{K}_2^{\text{II}}(a) = \emptyset$. According to the discussion in Section \ref{S:lower_level_raw}, the following results can be expected.

\begin{itemize}
\item For $k \in \mathcal{K}_1(a)$, the term $\left(a h_k  - \beta_k\right)^2 = \left(a h_k \cdot \frac{\beta_k^{\max}}{a h_k}  - \beta_k^{\max}\right)^2 = 0$.
\item For $k \in \mathcal{K}_2^{\text{I}}(a)$, the term $\left(a h_k b_k - \beta_k\right)^2 = \left(a h_k b_k^{\max} - \beta_k^{\max}\right)^2$.
\item For $k \in \mathcal{K}_2^{\text{II}}(a)$, the term $\left(a h_k b_k - \beta_k\right)^2 = \frac{\left(\lambda^*(a)\right)^2}{4 {c_k^2}}$, which is from (\ref{e:beta_lambda_express}).
\end{itemize}
Collecting the above results,
the cost function of Problem \ref{p:upper_level_raw} can be rewritten as
\begin{equation} \label{e:E_a_reformulate}
E(a) = \sum_{k\in \mathcal{K}_2^{\text{I}}(a)} \left(a h_k b_k^{\max} - \beta_k^{\max}\right)^2 + \sum_{k\in \mathcal{K}_2^{\text{II}}(a)}\frac{\left(\lambda^*(a)\right)^2}{4 {c_k^2}} + a^2 \sigma^2
\end{equation}
whose convexity with $a$ is not straightforward to see.

To explore the convexity of $E(a)$ with $a$, we first investigate the convexity of the term $\frac{\left(\lambda^*(a)\right)^2}{4 {c_k^2}}$ with $a$.
For the ease of presentation in the following, define $F(a) = - \frac{\lambda^*(a)}{2 {c_k}}$, then we have $F(a)>0$ since $\lambda^*(a)<0$ and the term $\frac{\left(\lambda^*(a)\right)^2}{4 {c_k^2}}$ can be written as $(F(a))^2$.
Moreover, by defining the function
\begin{equation}
P(a, x) \triangleq \sum_{k \in \mathcal{K}_1(a)} \beta_k^{\max} + \sum_{k \in \mathcal{K}_2(a)} \min \left(a h_k b_k^{\max} + x, \beta_k^{\max}\right),
\end{equation}
it can be found that the $F(a)$ is actually the solution of the equation
$P(a, x) = 1$ according to (\ref{e:beta_lambda_sum_1}).
Considering the fact that the function $P(a, x)$ is a non-decreasing function with $x$, the $F(a)$ can be expressed in another way, which is given as follows.
\begin{prob} \label{p:F_a_def}
\begin{subequations}
\begin{align}
F(a) \triangleq \mathop{\min} \limits_{x} \quad & x \nonumber \\
\text{s.t.} \quad & P(a, x) \geq 1, \\
               & x \geq 0.
\end{align}
\end{subequations}
\end{prob}
For the function $F(a)$, the following property can be expected.
\begin{lem} \label{lem:F_monotonicity}
The function $F(a)$ is non-increasing function with $a$ when $a$ is in a region in which $\mathcal{K}_1(a)$ and $\mathcal{K}_2(a)$ do not change.
\end{lem}
\begin{IEEEproof}
Please refer to Appendix \ref{app:F_monotonicity}.
\end{IEEEproof}
\begin{lem} \label{lem:F_convexity}
The function $F(a)$ is convex with $a$ when $a$ is in a region in which $\mathcal{K}_1(a)$ and $\mathcal{K}_2(a)$ do not change.
\end{lem}
\begin{IEEEproof}
Please refer to proof in Appendix \ref{app:F_convexity}.
\end{IEEEproof}
\begin{remark} \label{r:F_convexity}
Based on Lemma \ref{lem:F_convexity}, when $a$ is in a region in which $\mathcal{K}_1(a)$ and $\mathcal{K}_2(a)$ do not change, the function $(F(a))^2$ is also a convex function with $a$ since it is a composition of a convex function $F(a)$ with a non-decreasing convex function $z(x)=x^2 \left(\text{when~} x \geq 0 \right)$, which is still convex according to \cite{boyd_vandenberghe_2004}.
\end{remark}

According to Remark \ref{r:F_convexity}, the cost function of Problem \ref{p:upper_level_raw}, i.e., $E(a)$ in (\ref{e:E_a_reformulate}), becomes a convex function with $a$, when the sets $\mathcal{K}_1(a)$, $\mathcal{K}_2^{\text{I}}(a)$, and $\mathcal{K}_2^{\text{II}}(a)$ are unchanged.
When $a$ varies in a $\mathcal{B}_{m}$ interval ($m=0, 1, 2, ..., M$), both the set $\mathcal{K}_1(a)$ and the set $\mathcal{K}_2(a)$ keep unchanged, while the sets $\mathcal{K}_2^{\text{I}}(a)$ and $\mathcal{K}_2^{\text{II}}(a)$ may vary. Next, we divide interval $\mathcal{B}_{m}$ into a number of sub-intervals, and in each of the sub-intervals, $\mathcal{K}_2^{\text{I}}(a)$ and $\mathcal{K}_2^{\text{II}}(a)$ do not change.

According to (\ref{e:beta_lambda_express}), for $k \in \mathcal{K}_2(a)$, it can be found that:
1) When $F(a) \geq \left(\beta_k^{\max} - a h_k b_k^{\max}\right)$, we have $k \in \mathcal{K}_2^{\text{I}}(a)$;
2) When $F(a) < \left(\beta_k^{\max} - a h_k b_k^{\max}\right)$, we have $k \in \mathcal{K}_2^{\text{II}}(a)$. Note that $F(a)$ is a non-increasing convex function, and $\left(\beta_k^{\max} - a h_k b_k^{\max}\right)$ is a linear decreasing function for $a \in \mathcal{B}_{m}$.
{
Thus, for the $k$th mobile device, the function $F(a)$ and the line $\left(\beta_k^{\max} - a h_k b_k^{\max}\right)$ may have no intersection (when $F(a)$ is always above $\left(\beta_k^{\max} - a h_k b_k^{\max}\right)$), one intersection (when $\left(\beta_k^{\max} - a h_k b_k^{\max}\right)$ is tangent with $F(a)$ at some point), two intersections, which corresponds to one, two, and three sub-intervals such that the $k$th mobile device is always in $\mathcal{K}_2^{\text{I}}(a)$ or always in $\mathcal{K}_2^{\text{II}}(a)$.
}
Recall that for $a \in \mathcal{B}_{m}$, $\mathcal{K}_2(a)$ contains $|\mathcal{K}_2(a)| = K-m$ mobile devices. So interval $\mathcal{B}_{m}$ can be divided into at most $2(K-m)+1$ sub-intervals such that when $a$ is in one sub-interval, the set $\mathcal{K}_2^{\text{I}}(a)$ and set $\mathcal{K}_2^{\text{II}}(a)$ keep unchanged.
{
The $2(K-m) + 1$ sub-intervals is calculated in such a way: For each mobile device belonging to the set $\mathcal{K}_2$, there are at most two intersections. For a total number of $K-m$ mobile devices, there are at most $2\left(K-m\right)$ intersections, which could divide the region $\mathcal{B}_m$ into $2(K-m)+1$ sub-intervals.
}
Thus, in each sub-interval of $\mathcal{B}_{m}$, the cost function of Problem \ref{p:upper_level_raw}, i.e., $E(a)$ in (\ref{e:E_a_reformulate}), is a convex function with $a$, and the optimal $a$ in the sub-interval can be found by a Golden search method.

Then the optimal value of $a$ for $a\in \mathcal{A}_m$ ($m=0,1,...,M$) can be found accordingly. Then the optimal value of $a$ for $a\in (0,\infty)$ can be found by comparing the minimal cost function in intervals $\mathcal{A}_0$, $\mathcal{A}_1$,... $\mathcal{A}_{M+1}$.

Complexity Analysis: To find the optimal value of $a$, the major computation complexity is on the Golden search method for every sub-interval of $\mathcal{B}_m$ ($m=0, 1, ..., M$), with total worst-case complexity being $O(K^2)$ for all sub-intervals of $\mathcal{B}_0, \mathcal{B}_1,...,\mathcal{B}_M$.

{
\section{Convergence and Optimality for Training} \label{s:convergence}
}

{
In terms of convergence and optimality, \cite{wang2021federated,cao2021optimized,journals/jsac/CaoZXC22} have already investigated this topic for a general over-the-air computation aided FL system.
According to their results, our proposed method can achieve convergence or even zero optimality gap, under some mild conditions.
It is worth mentioning that the derived optimality gap is highly related to the MSE of aggregation, which also necessities MSE minimization as we do in this paper.
}

{
To be specific, for a non-convex loss function being $L$-smooth (whose definition can be found in \cite{wang2021federated,cao2021optimized,journals/jsac/CaoZXC22}), which is applicable for the training of a neural network, the associated results on convergence or optimality gap in \cite{wang2021federated,cao2021optimized,journals/jsac/CaoZXC22} are given as follows:
\begin{enumerate}
     \item {\cite{wang2021federated} investigates an intelligent reflecting surfaces (IRS) enhanced over-the-air computation aided FL system, which also applies for our considered system since the IRS system can also realize any channel gain between mobile devices and the BS assumed in this paper. With a general setup of signal amplification factors at both mobile devices and the BS,
the average norm of the cost function's gradient over $T$ iterations is upper bounded as
    \begin{equation}\label{gap-4}
        \begin{aligned}
            &\frac{1}{T} \sum_{t=0}^{T-1} \mathbb{E} \left\{ \left\lVert \nabla f(\bm{w}_t) \right\rVert^2 \right\}
            \le O \left({\frac{1}{T^2} \sum_{t=0}^{T-1}} \mathbb{E} \left\{\left\lVert \varepsilon_t \right\rVert^2 \right\} \right) + O \left( \frac{1}{T} \left( f\left(\bm{w}_0\right) - f^* \right)  \right),
        \end{aligned}
    \end{equation}
where $\varepsilon_t$ denotes the aggregation error, $\mathbb{E} \left\{\left\lVert \varepsilon_t \right\rVert^2 \right\} $ is the MSE of aggregation at the $t-$th round of iteration, $\bm{w}_0$ and $\bm{w}_t$ represent the global parameter vector at the beginning of training and the $t-$th round of iteration respectively,
$f^*$ is the minimal achievable value of the loss function $f(\cdot)$.
It can be observed that for a fixed $T$,  the averaged norm of loss function's gradient is highly related to the MSE of aggregation in each step of iteration.
Moreover, as $T \rightarrow \infty$, the averaged norm of loss functions's gradient trends to be zero.
This verifies the convergence of our proposed method.
   }
\item {With $\tau$-Polyak-Łojasiewicz condition (whose definition can be found in \cite{cao2021optimized,journals/jsac/CaoZXC22}),
 according to \cite{cao2021optimized,journals/jsac/CaoZXC22}, the optimality gap can be written as
    \begin{equation}\label{gap-3}
        \begin{aligned}
            &\mathbb{E}\{f\left(\bm{w}_T\right)  \} - f^* \le O \left({\frac{1}{T}} \mathbb{E} \left\{ \left\lVert \varepsilon_t \right\rVert^2 \right\} \right) + O \left({\frac{1}{T}} \left(f\left(\bm{w}_0\right) - f^* \right)  \right) + {\frac{1}{T}} W,
        \end{aligned}
    \end{equation}
    where $W$ is a constant item related to system parameters such as the $L$, $\tau$, and the upper bound of local gradient's variance.
    This result can verify not only the convergence but also the ability of achieving zero optimality gap for our proposed method.
    }
\end{enumerate}
}

\section{Numerical Results} \label{s:numerical_results}

In this section, numerical results are presented to evaluate the performance of our proposed method. Default system settings are given as follows. There are $K=20$ mobile devices in the FL system \cite{shi2020joint}. The $D_k$ for $k=1, 2, ..., K$ are randomly generated to be 3979, 3974, 3985, 3933, 4026, 3984, 3972, 3961, 3991, 3986, 4051, 3972, 3921, 3991, 3983, 3937, 3958, 4058, 4033, and 4051, respectively.
{
$c_k$ is selected to be 1 for $\forall k\in \mathcal{K}$.}
$S_T$ is selected to be $40000$.

{
We compare the performance of our proposed method with the AirFedSGD method \cite{journals/icl/ZhangTWS23}, the Transmission Power Control (referred to as the TPC method in the sequel) in \cite{cao2020optimized}, and the Computation Optimal Policy (referred to as the COP method in the sequel) in \cite{liu2020over}.}
}
{
The AirFedSGD method is the traditional FedSGD method \cite{conf/aistats/McMahanMRHA17} adapted to the environment of over-the-air computation, which sets the signal amplification factor $b_k$ to be inversely proportional to the channel gain for every mobile device $k \in \mathcal{K}$.
The amplification factor of the received signal at the BS $a$ is simply set as $\frac{1}{\left\lvert \mathcal{K} \right\rvert }$.
The TPC method follows the idea of AirFedSGD but considers the case that the channel condition may be poor. In this case, the associated signal amplification factor has to be very high by following AirFedSGD method but cannot be achieved due to the limit of mobile device's transmit power. The TPC method regulates the signal amplification factor to be at its maximal transmit power level for this case. The amplification factor of the received signal at the BS $a$ is further optimized so as to minimize the aggregation MSE, which generates a closed-form solution.
The COP method optimizes the signal amplification factor $b_k$ for $k\in \mathcal{K}$ and the amplification factor of the received signal at the BS $a$ jointly to minimize the aggregation MSE.
For the above three methods, they all select $S_k$ as $D_k$ for $k\in \mathcal{K}$.
}

{For ease of comparison with the AirFedSGD, the TPC, and the COP method, we also adopt their parameter setup \cite{liu2020over,cao2020optimized} of $\sigma^2$, $h_k$, and $b_k^{\max}$ for $k\in \mathcal{K}$ as follows}: $\sigma^2=1$; the channel coefficient $h_k$ for $k\in \mathcal{K}$ are i.i.d Rayleigh distributed random variables with mean being $\bar{h}=\sqrt{{\pi}/{\left(4-\pi\right)}}$ (this implies a variance being 1);  $b_k^{\max}$ is set as $\sqrt{10}$ for $k\in \mathcal{K}$. {The code to realize our numerical results is available at https://github.com/anxuming/FedAirComp.git.}

\begin{figure}
  \centering
  \begin{minipage}{0.49\linewidth}
    \centering
    \includegraphics[width=0.7 \figwidth]{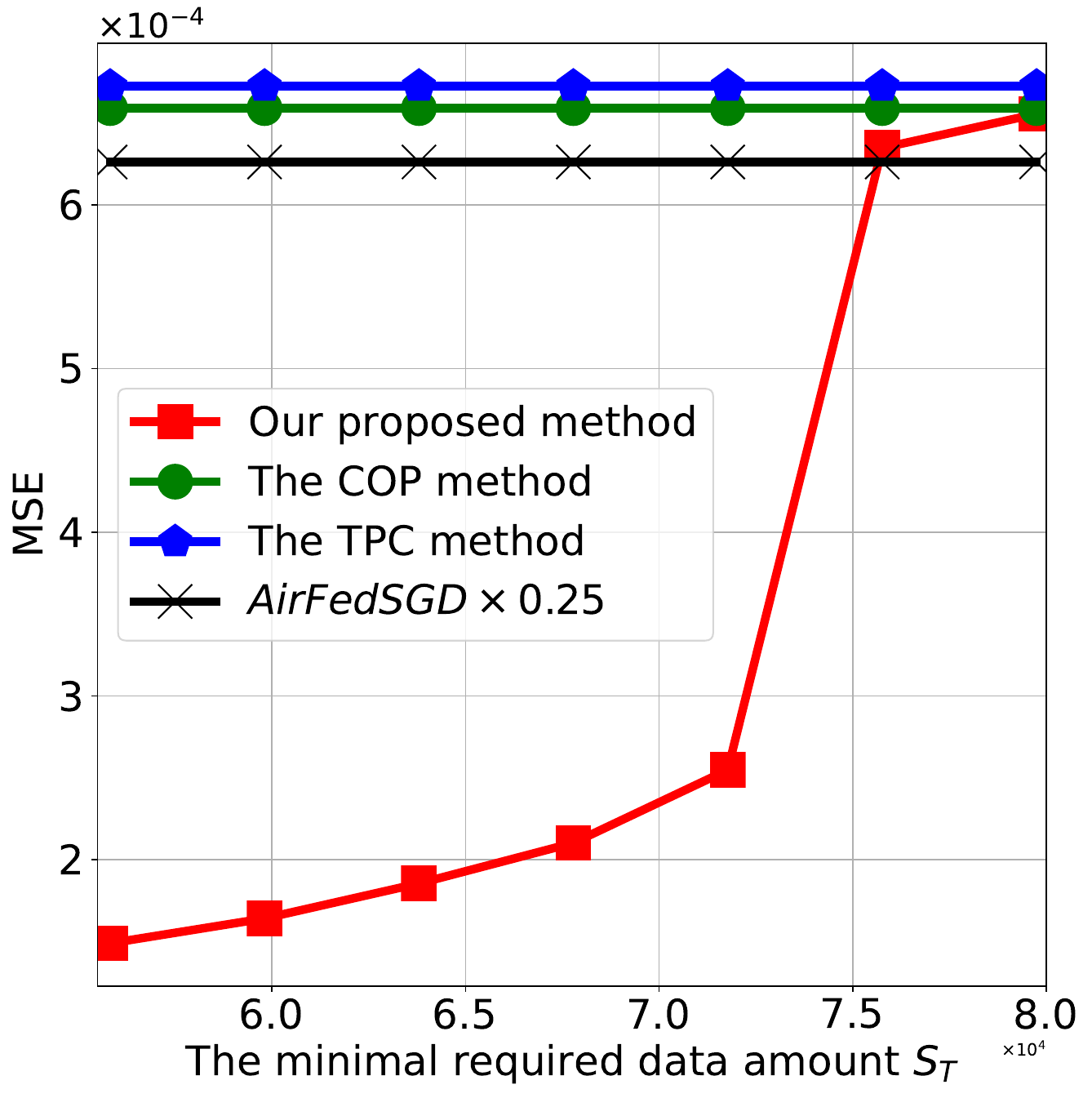}
    \caption{MSE versus minimal required data amount $S_T$.}
    \label{data-re}
  \end{minipage}
  \begin{minipage}{0.49\linewidth}
    \centering
    \includegraphics[width=0.7 \figwidth]{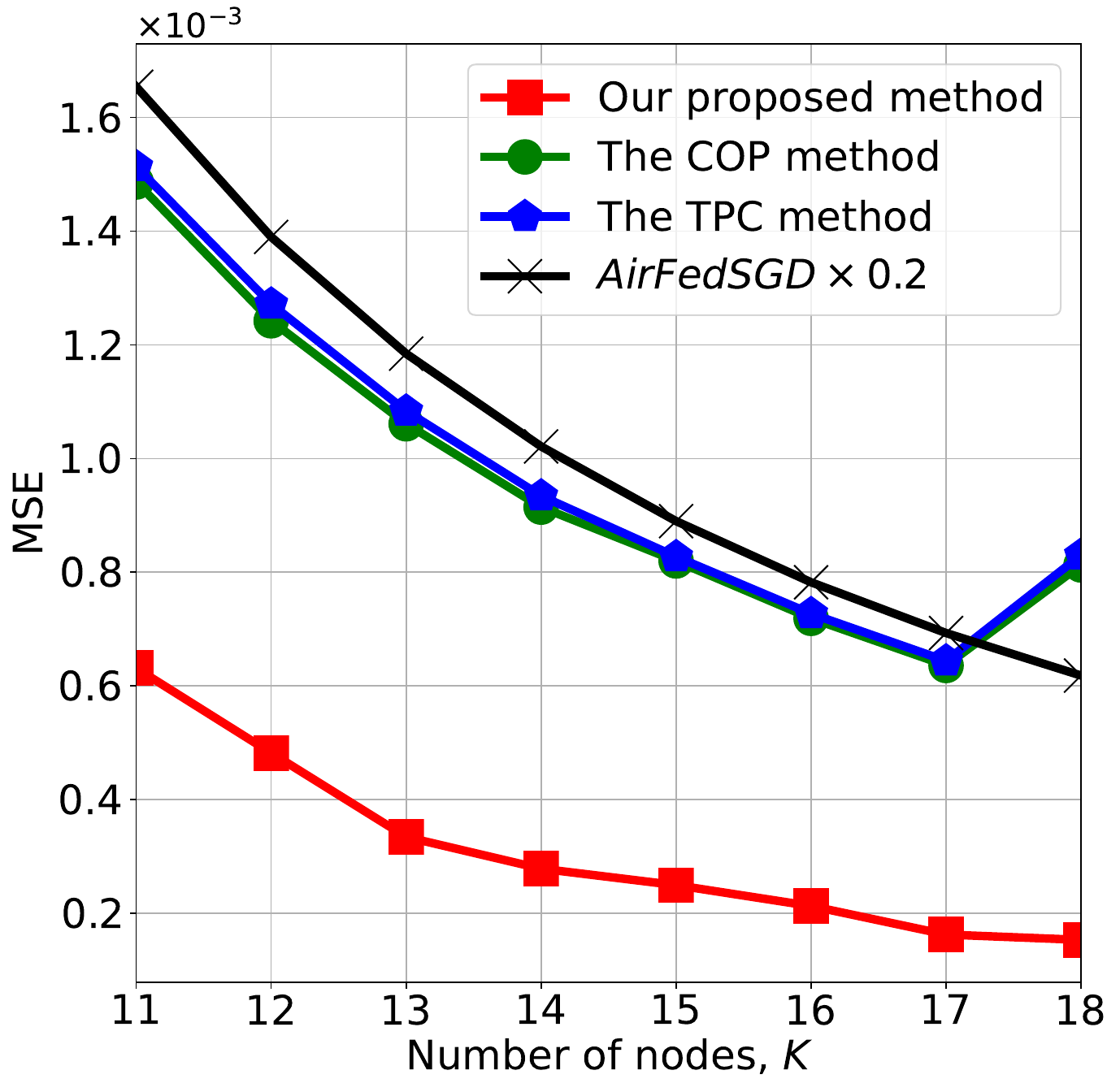}
    \caption{MSE versus the number of mobile devices $K$.}
    \label{node-num}
  \end{minipage}

  \begin{minipage}{0.49\linewidth}
    \centering
    \includegraphics[width=0.7 \figwidth]{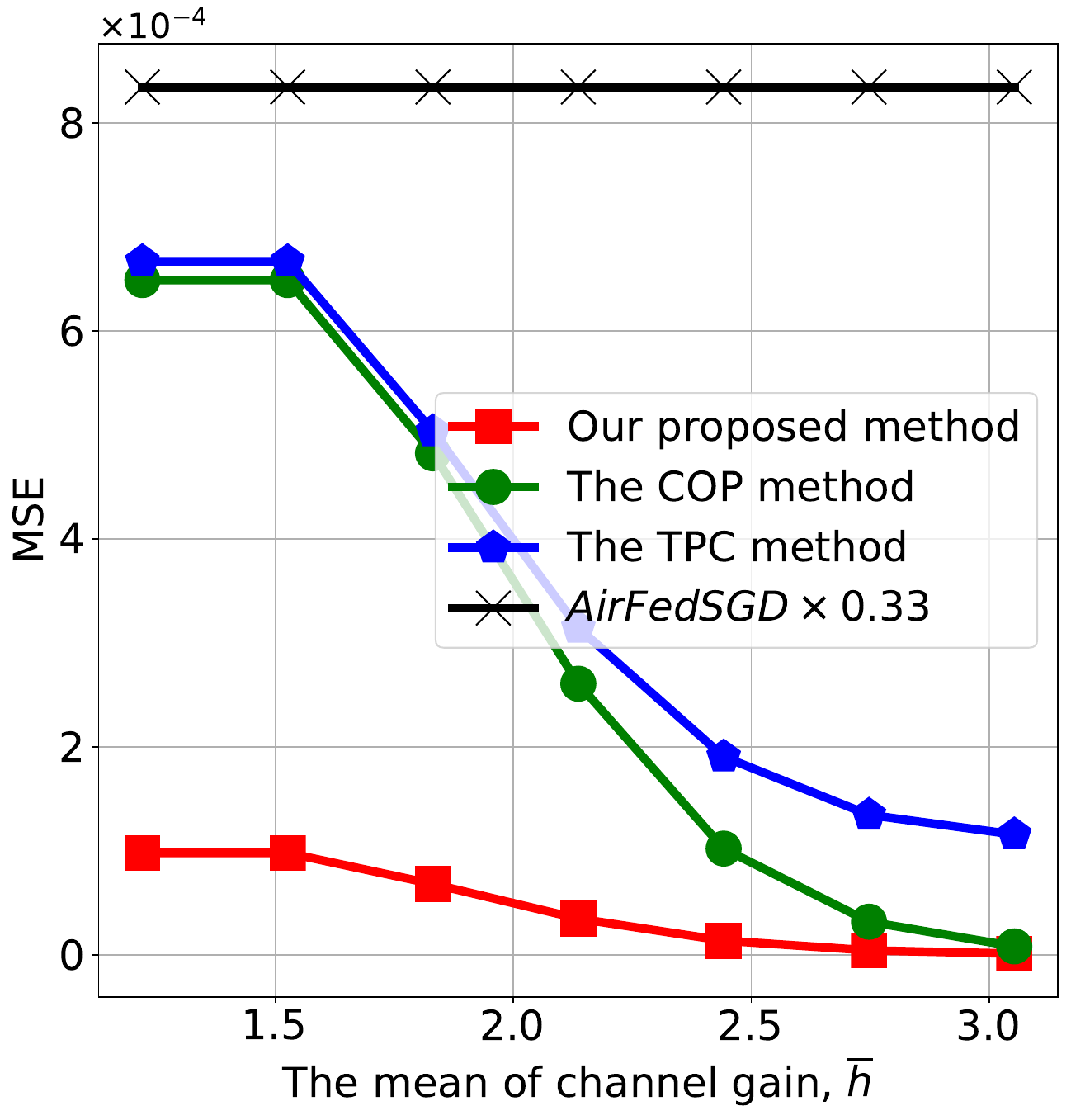}
    \caption{MSE versus the mean of channel coefficient $\bar{h}$.}
    \label{channel}
  \end{minipage}
  \begin{minipage}{0.49\linewidth}
    \centering
    \includegraphics[width=0.7 \figwidth]{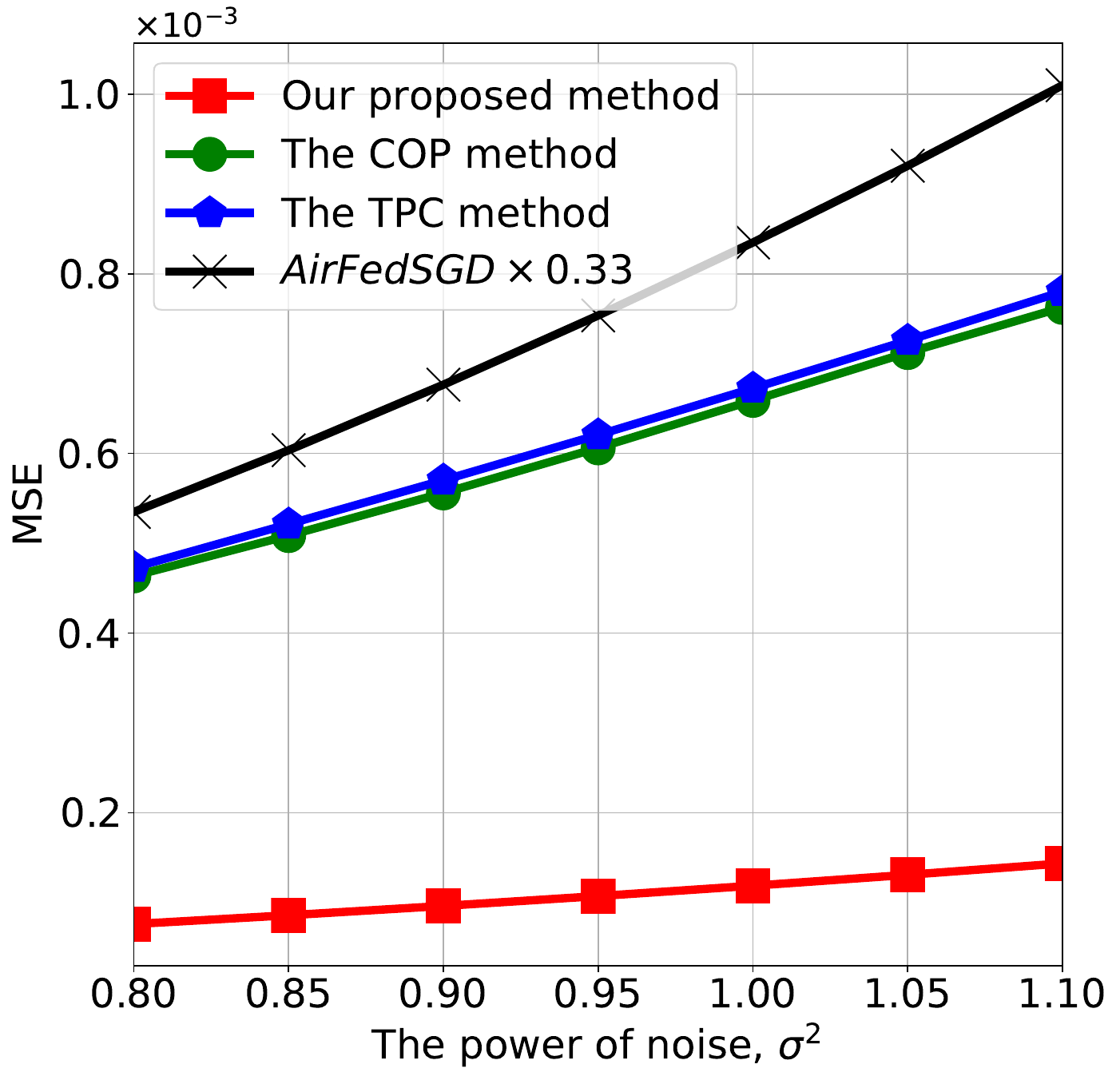}
    \caption{MSE versus the variance of noise $\sigma^2$.}
    \label{noise}
  \end{minipage}
\end{figure}

\subsection{MSE Improvement}

{
Fig. \ref{data-re}, Fig. \ref{node-num}, Fig. \ref{channel}, and Fig. \ref{noise} plot the MSE versus various system parameters for our proposed method, the AirFedSGD method, the TPC method, and the COP method.
It can be seen that our proposed method can always achieve an MSE no larger than the one under the COP method.
Moreover, the COP method also achieves a lower MSE than the ones under the AirFedSGD method and the TPC method.
These results demonstrates the effectiveness of our proposed method.
The reason behind can be explained as follows.
Our proposed method minimizes the MSE through optimizing one more dimension of variable $S_k$ for $k\in \mathcal{K}$, while the COP method simply sets $S_k$ as $D_k$ for $k\in \mathcal{K}$.
Compared with the COP method, both the AirFedSGD method and the TPC method offer a sub optimal solution for minimizing the MSE also with $S_k$ being set as $D_k$ for $k\in \mathcal{K}$. Hence they achieve higher MSE than the COP method.
Since the AirFedSGD method and the TPC method are always inferior to the COP method, we will mainly analyze the performance of our proposed method and the COP method in the following.
}

Fig. \ref{data-re} shows the MSE versus the minimal required data amount $S_T$ for our proposed method, {the AirFedSGD method, the TPC method}, and the COP method.
In the COP method, each mobile device utilizes all its data in its local training, and thus, its MSE does not change with $S_T$. In our proposed method, the MSE increases as $S_T$ grows. This is because the growing $S_T$ means that the feasible region of Problem \ref{p:raw_opt} shrinks, and thus, its cost function (i.e., the MSE) increases. When $S_T=\sum_{k=1}^{K}{D_K}$, the MSE of our proposed method and the COP method are the same. This is because, to satisfy $S_T=\sum_{k=1}^{K}{D_K}$,  each mobile device in our proposed method has to use all its data in its local training.

Fig. \ref{node-num} shows the MSE versus the number of mobile devices $K$. When a number of $K$ mobile devices are investigated, we select the first $K$ mobile devices from the set of 20 mobile devices described at the beginning of this section.
From Fig. \ref{node-num}, it can be seen that the MSE under the CoP method first goes down with $K$ and then increase after $K=17$. The reason is as follows. When the number of mobile devices increases,  $\beta_k$ of the $k$th mobile device tends to decrease (recalling that the summation of $\beta_k$'s is equal to one). Thus, the $k$th mobile device can have a smaller term $(a h_k b_k - \beta_k)^2$ in the expression of MSE, leading to a decreasing trend of MSE. On the other hand, the increase of the number of mobile devices means that there are more terms $(a h_k b_k - \beta_k)^2$ in the expression of MSE, leading to an increasing trend of MSE. When $K$ is smaller than 17, the decreasing trend dominates. When $K$ increases beyond 17, the increasing trend dominates.

From Fig. \ref{node-num}, it can also be seen that the MSE under our proposed method decreases as $K$ grows. The reason is as follows. Define the optimal solution of $b_k$ and $\beta_k$ as $b_k^{K}$ and $\beta_k^{K}$ for $k = 1, 2, ..., K$, respectively, when solving Problem \ref{p:lower_level_raw} with $K$ mobile devices. Accordingly, the minimal cost function of Problem \ref{p:lower_level_raw}, i.e., the minimal MSE with given $a$, can be expressed as $E_{K}(a)$. Then for any $a$, we have
  \begin{equation} \label{e:simu}
    \begin{aligned}
      E_{K}(a) =& \sum_{k=1}^{K} \left(ab_k^{K} h_k-\beta_k^{K}\right)^2 c_k + a^2\sigma^2 \\
      =&  \sum_{k=1}^{K} {\left(ab_k^{K}h_k-\beta_k^{K}\right)^2} c_k + \left(a\cdot 0\cdot h_{K+1}-0\right)^2  c_{K+1} + a^2\sigma^2  \\
      \ge& \sum_{k=1}^{K+1} {\left(ab_k^{K+1}h_k-\beta_k^{K+1}\right)^2} c_k + a^2\sigma^2
      = E_{K+1}(a)
    \end{aligned}
  \end{equation}
where the inequality in (\ref{e:simu}) holds since $\{b_1^{K}, ..., b_{K}^{K}, 0\}$ and $\{\beta_1^{K}, ..., \beta_{K}^{K}, 0\}$ is also a feasible solution of $\{b_k\}$ and $\{\beta_k\}$ for Problem \ref{p:lower_level_raw} with $K+1$ mobile devices.
Then for the upper-level problem, i.e., Problem \ref{p:upper_level_raw}, denoting the optimal solution of $a$ as $a_{K}$ with $K$ mobile devices, we have
\begin{equation}\nonumber
  \begin{aligned}
    E_{K}(a_{K}) \ge E_{K+1}(a_{K}) \ge E_{K+1}(a_{K+1}),
  \end{aligned}
\end{equation}
which means that the MSE decreases as $K$ grows for our proposed method.

In Fig. \ref{channel}, the MSE is plotted versus the mean of channel coefficient $\bar{h}$. It can be observed that the MSE decreases with $\bar{h}$ for all the methods. This can be explained as follows.
With the increase of $\bar{h}$, $h_k$ for $k\in \mathcal{K}$ tends to be larger, and thus, more mobile devices are included in set $\mathcal{K}_1$, and fewer mobile devices are included in set $\mathcal{K}_2$. So in our cost function $\sum_{k=1}^{K} \left(a b_k h_k  - \beta_k\right)^2 + a^2 \sigma^2$, we will have more terms of $\left(a b_k h_k - \beta_k \right)^2$ equal to zero (recalling that $\left(a b_k h_k - \beta_k \right)^2=0$ for $k\in \mathcal{K}_1$). For the terms of $\left(a b_k h_k - \beta_k \right)^2$ for $k\in \mathcal{K}_2$,
although they cannot be equal to be zero, they have a higher chance to get a smaller value thanks to the increase of $h_k$. Thus, the MSE of our proposed method decreases. Due to a similar reason, the MSE of the COP method
also decreases, recalling the fact that the COP method is actually to solve our Problem \ref{p:raw_opt}
by setting $S_k$ as $D_k$ for $k\in \mathcal{K}$.

In Fig. \ref{noise}, the MSE is plotted versus the variance of noise $\sigma^2$. It can be seen that the MSE increases as $\sigma^2$ grows for both our proposed method and the COP method. For our proposed method, the reason can be explained as follows. Define the cost function of Problem \ref{p:raw_opt} as $E\left(a, \{b_k\}, \{S_k\}, \sigma^2\right)$, and the optimal solution of $a$, $\{b_k\}$, and $\{S_k\}$ for solving Problem \ref{p:raw_opt} (when the variance of noise is $\sigma^2$) are given as $a^*(\sigma^2)$, $\{b_k^*(\sigma^2)\}$, and $\{S_k^*(\sigma^2)\}$, respectively. Then the minimal achievable cost function of Problem \ref{p:raw_opt} can be written as
$E\left(a^*(\sigma^2), \{b_k^*(\sigma^2)\}, \{S_k^*(\sigma^2)\}, \sigma^2\right)$.
For $\sigma_1^2 < \sigma_2^2$, {we have}
\begin{equation}
\begin{split}
& E\left(a^*(\sigma_1^2), \{b_k^*(\sigma_1^2)\}, \{S_k^*(\sigma_1^2)\}, \sigma_1^2\right) \\
\leq & E\left(a^*(\sigma_2^2), \{b_k^*(\sigma_2^2)\}, \{S_k^*(\sigma_2^2)\}, \sigma_1^2\right) \\
\leq & E\left(a^*(\sigma_2^2), \{b_k^*(\sigma_2^2)\}, \{S_k^*(\sigma_2^2)\}, \sigma_2^2\right)
\end{split}
\end{equation}
in which the second inequality is because $E\left(a^*(\sigma^2), \{b_k^*(\sigma^2)\}, \{S_k^*(\sigma^2)\}, \sigma^2\right)$ is an increasing function of $\sigma^2$.
As of the curve for the COP method, it can be explained in the same way.

\subsection{Training Performance Improvement}
In this sub-section, the performance of training a neural network ResNet-18 through our proposed method is analyzed and compared with benchmark methods.

\begin{figure}
  \centering
  \begin{minipage}{0.49\linewidth}
    \centering
    \includegraphics[width=1.0\figwidth]{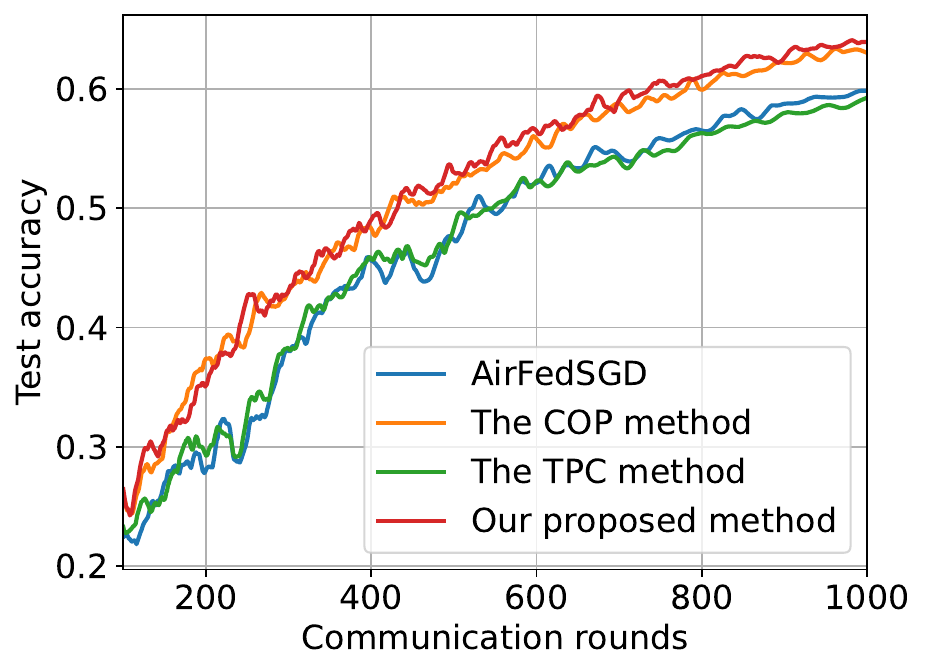}
    \caption{Test accuracy on CIFAR10.}
    \label{c10-test}
  \end{minipage}
  \begin{minipage}{0.49\linewidth}
    \centering
    \includegraphics[width=1.0\figwidth]{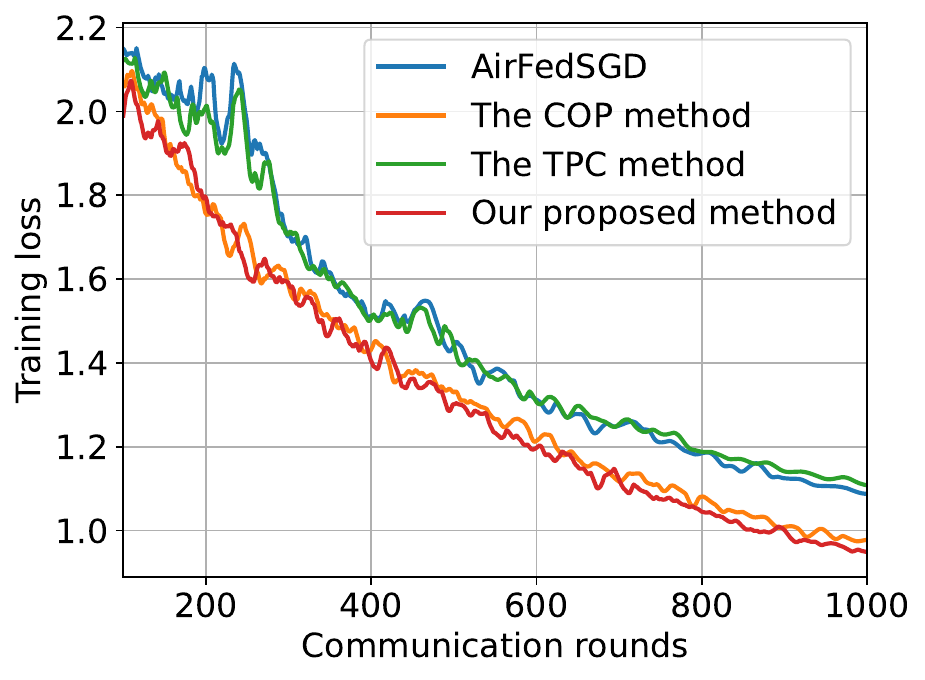}
    \caption{Training loss on CIFAR10}
    \label{c10-train}
  \end{minipage}
\end{figure}

{Fig. \ref{c10-test} and Fig. \ref{c10-train} present the performance of training ResNet-18 based on CIFAR10 dataset by utilizing our proposed method and benchmark methods.
It can be observed that our proposed method trends to outperform the benchmark methods as the communication round $T$ grows, in terms of not only test accuracy but also training loss.
This verifies that our effort on reducing the MSE for gradient aggregation can help to improve training performance.
}

\begin{figure}
  \centering
  \begin{minipage}{0.49\linewidth}
    \centering
    \includegraphics[width=1.0\figwidth]{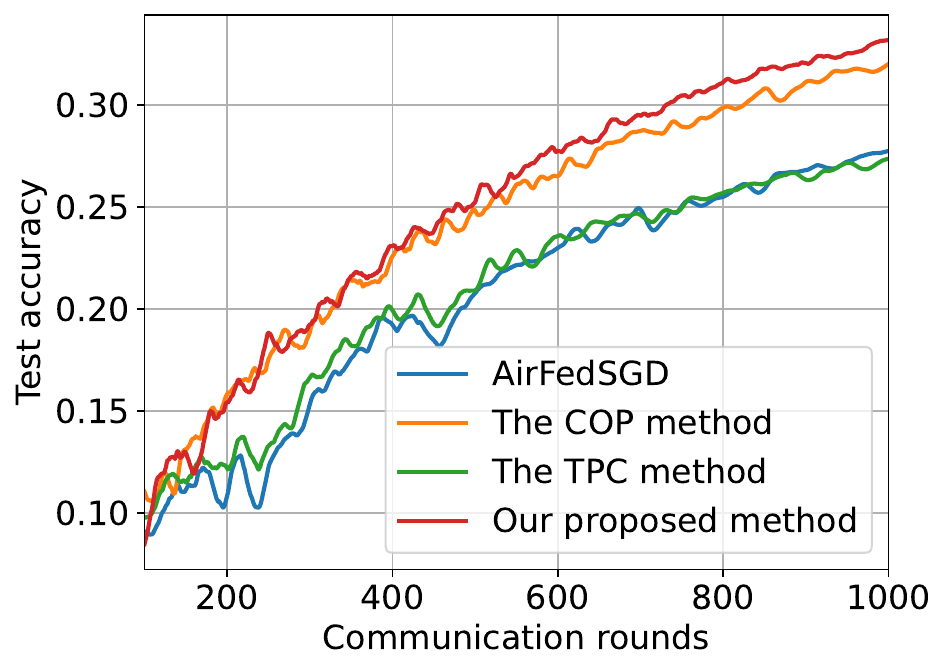}
    \caption{Test accuracy on CIFAR100.}
    \label{c100-test}
  \end{minipage}
  \begin{minipage}{0.49\linewidth}
    \centering
    \includegraphics[width=1.0\figwidth]{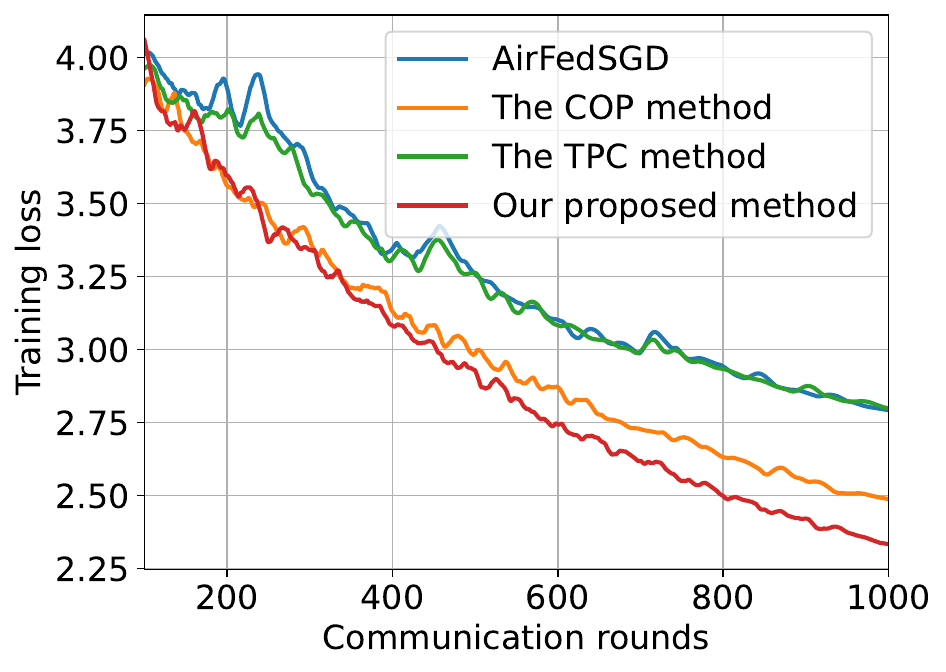}
    \caption{Training loss on CIFAR100.}
    \label{c100-train}
  \end{minipage}
\end{figure}

{In Fig. \ref{c100-test} and Fig. \ref{c100-train}, the performance comparison for training ResNet-18 is evaluated on another dataset CIFAR100. Results similar to Fig. \ref{c10-test} and \ref{c10-train} can be also obtained. This strengthens the meaningfulness of our efforts on reducing the MSE for gradient aggregation.
}

\begin{figure}
  \centering
  \begin{minipage}{0.49\linewidth}
    \centering
    \includegraphics[width=1.0\figwidth]{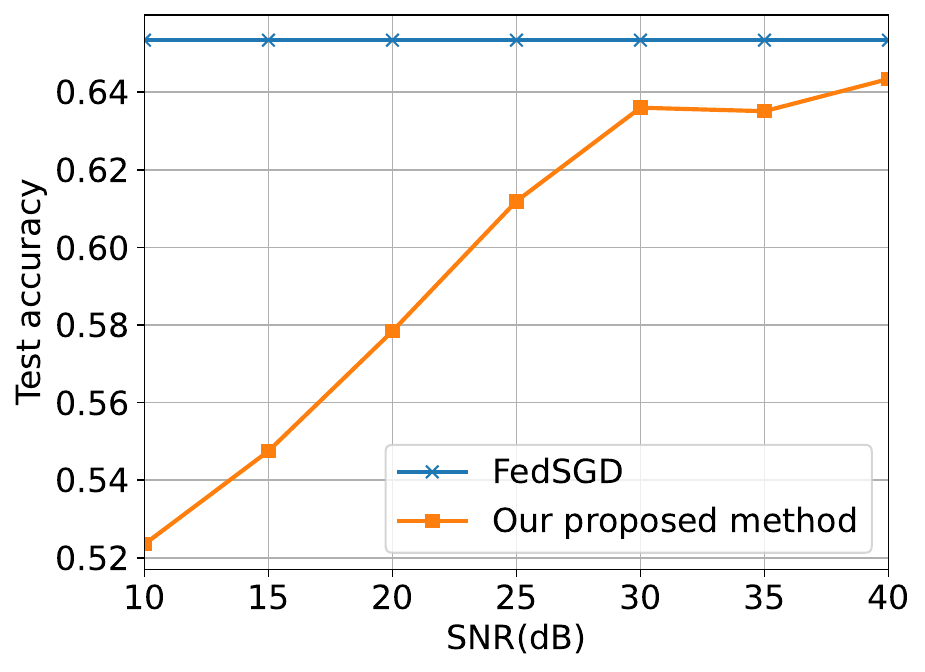}
    \caption{Test accuracy on CIFAR10 with different SNR.}
    \label{c10-snr-test}
  \end{minipage}
  \begin{minipage}{0.49\linewidth}
    \centering
    \includegraphics[width=1.0\figwidth]{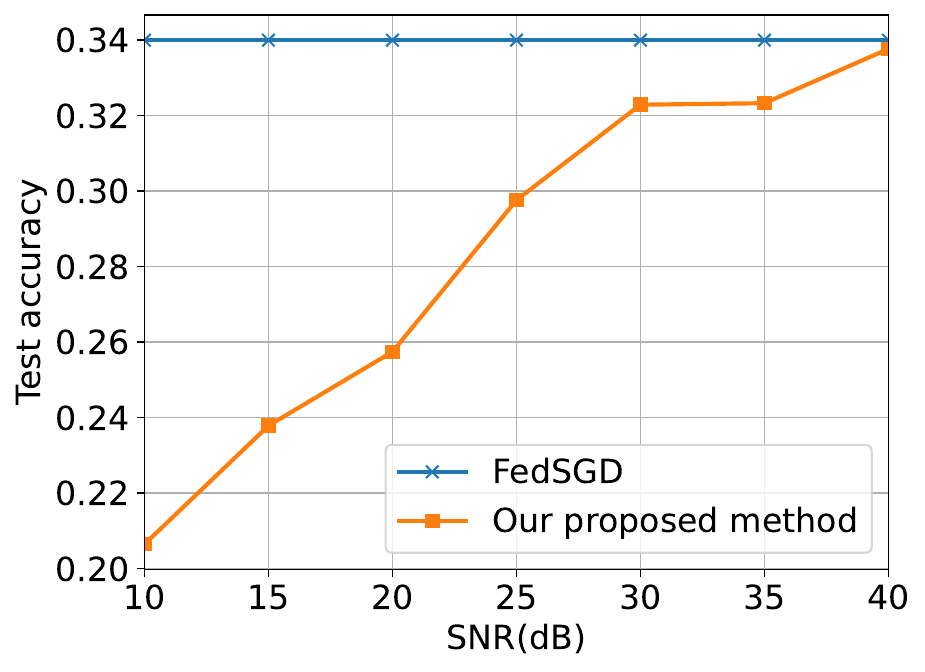}
    \caption{Test accuracy on CIFAR100 with different SNR.}
    \label{c100-snr-test}
  \end{minipage}
\end{figure}

{
Define ${\left(b_k^{\max}\right)^2}/{\sigma^2}$ as the signal-to-noise ratio (SNR) for $k\in \mathcal{K}$, in Fig. \ref{c10-snr-test} and Fig. \ref{c100-snr-test}, we plot how the test accuracy varies with the SNR for our proposed method, under CIFAR10 and CIFAR100 datasets respectively.
As a comparison, the performance of the traditional FedSGD method \cite{conf/aistats/McMahanMRHA17}, which is noise free for aggregation, is also plotted.
From Fig. \ref{c10-snr-test} and Fig. \ref{c100-snr-test}, it can be observed that as the SNR goes up, the test accuracy of our proposed method trends to increase and approaches to the performance of the FedSGD method.
These results offer such an inspiration for improving the test accuracy of our proposed method: We can increase $b_k^{\max}$ for $k\in \mathcal{K}$ to overcome the negative effect of additive noise for aggregation to achieve a similar performance like the noise free aggregation method, such as the FedSGD.}

\begin{figure}
  \centering
  \begin{minipage}{0.49\linewidth}
    \centering
    \includegraphics[width=1.0\figwidth]{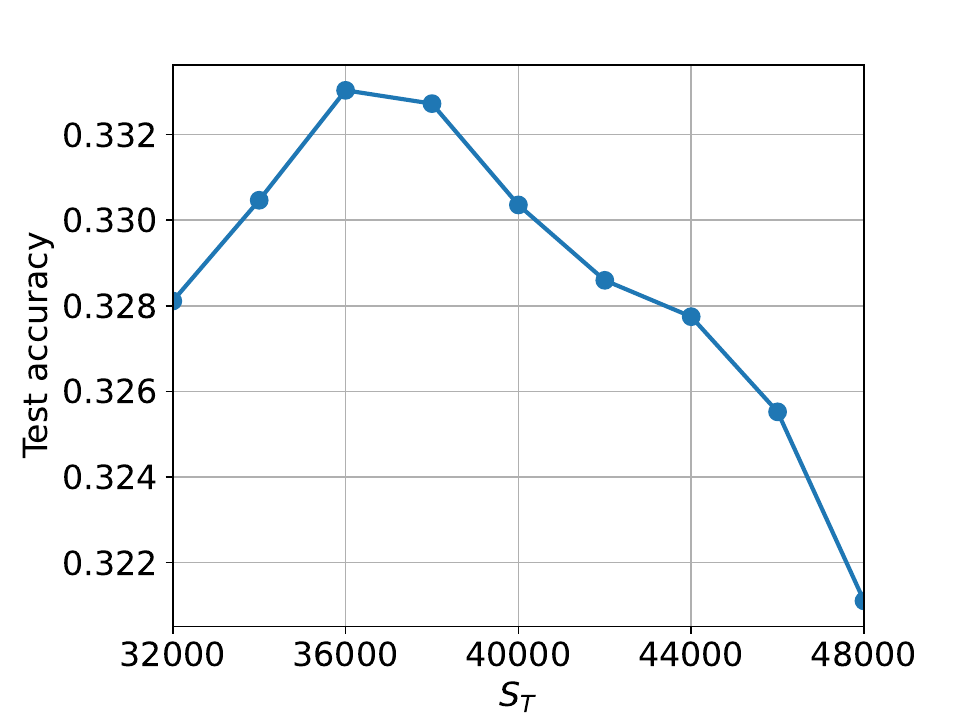}
    \caption{Test accuracy on CIFAR100 versus $S_T$.}
    \label{c100-test-ST}
  \end{minipage}
  \begin{minipage}{0.49\linewidth}
    \centering
    \includegraphics[width=1.0\figwidth]{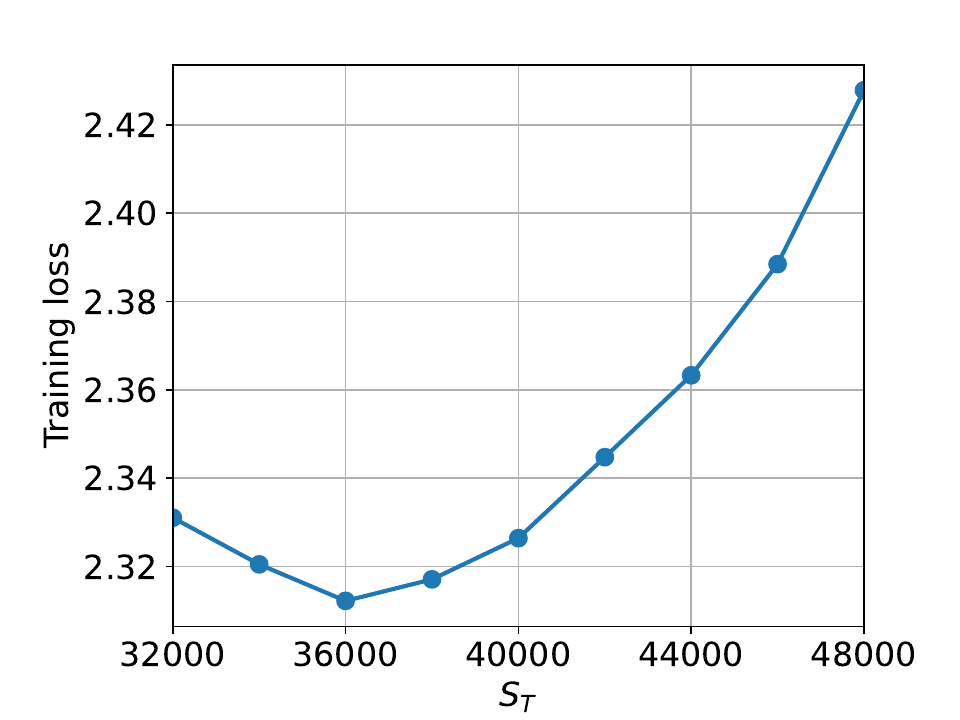}
    \caption{Training loss on CIFAR100 versus $S_T$.}
    \label{c100-train-ST}
  \end{minipage}
\end{figure}

{
In Fig. \ref{c100-test-ST} and Fig. \ref{c100-train-ST}, by utilizing our proposed method, the test accuracy and training loss for training ResNet-18 under dataset CIFAR100 are plotted versus $S_T$, respectively.
It can be seen that as $S_T$ grows, the performance of both test accuracy and training loss will first improve and then degrade.
The reason can be explained as follows.
When $S_T$ first grows, larger set of involving data samples is beneficial for improving the training performance.
As $S_T$ further increases, the negative effect of overfitting shows up, which leads to the degradation of training performance.
This result implies that it would not be optimal to use up all the available data samples, which backs up our operation on adjusting the data size for training.
Moreover, this result also suggests us to choose a proper $S_T$ value in real application.
}

\section{Conclusion and {Further Discussion}} \label{s:conclusion}
In this work, we have exploited data size selection for multiple mobile devices in a FL system powered by over-the-air computation.
The amplification factor at the mobile devices and the BS and the data sizes of the mobile devices are optimized jointly to minimize the MSE.
{
To solve the problem optimally, which is non-convex due to the coupling of multiple categories of variables and the existence of indicator function, we first simplify the cost function while preserving equivalence and perform variable transformation, and then solve the transformed problem in a two-level structure.}
Computation complexity of our proposed method is also analyzed, which is shown to be polynomial even in the worst case.
{
Numerical results illustrate that our proposed method can help to further reduce MSE and improve convergence performance compared with benchmark methods.
Our research results could provide helpful insights for the application of a FL system supported by over-the-air computation technique in the future.}
{
Moreover, recalling that the over-the-air computation technique can be also utilized for fusing the sensed data from multiple wireless sensors, our proposed method in this paper can be also extended to such a scenario. Specifically, when the mean value of some environmental parameter is to be estimated from a number of random observations at multiple wireless sensors, the number of random observations at each wireless sensor adopted for fusion, associated with the amplification factor at the wireless sensors  and the fusion center, can be adjusted to minimize the data fusion MSE with the help of our proposed method as well.
}

\begin{appendices}

\section{Proof of Lemma \ref{lem:indicator_remove}} \label{app:indicator_remove}
{
By defining $C_I(a, \{S_k\})$ as
\begin{prob} \label{p:raw_opt_orig_low}
\begin{subequations}
\begin{align}
 C_I(a, \{S_k\})  \triangleq
\mathop{\min} \limits_{\{b_k\}} & \quad \sum_{k=1}^{K}  \left(a b_k h_k  - \left(\frac{S_k}{\sum_{k=1}^{K} S_k}\right) \right)^2 {c_k {\mathcal{I}\left(S_k >0 \right)} }+ a^2 \sigma^2 \nonumber \\
\text{s.t.}   & \quad 0 \leq b_k \leq b_k^{\max}, \forall k \in \mathcal{K},
\end{align}
\end{subequations}
\end{prob}
it can be found that Problem \ref{p:raw_opt_orig} is equivalent with
\begin{prob} \label{p:raw_opt_orig_up}
\begin{subequations}
\begin{align}
\mathop{\min} \limits_{a, \{S_k\}} & \quad C_I(a, \{S_k\}) \nonumber \\
\text{s.t.}           & \quad 0 \leq S_k \leq D_k, \forall k \in \mathcal{K}, \label{e:S_k_interval}\\
                         & \quad \sum_{k=1}^K S_k \geq S_T, \label{e:S_k_sum}\\
                         & \quad a>0.
\end{align}
\end{subequations}
\end{prob}
}

{
By defining $C_{O}(a, \{S_k\})$ as
 \begin{prob} \label{p:raw_opt_low}
\begin{subequations}
\begin{align}
 C_{O}(a, \{S_k\})  \triangleq \mathop{\min} \limits_{\{b_k\}} & \quad \sum_{k=1}^{K}  \left(a b_k h_k  - \left(\frac{S_k}{\sum_{k=1}^{K} S_k}\right) \right)^2 { c_k }+ a^2 \sigma^2 \nonumber \\
\text{s.t.}   & \quad 0 \leq b_k \leq b_k^{\max}, \forall k \in \mathcal{K},
\end{align}
\end{subequations}
\end{prob}
then it can be found that Problem \ref{p:raw_opt} is equivalent with
\begin{prob} \label{p:raw_opt_up}
\begin{subequations}
\begin{align}
\mathop{\min} \limits_{a, \{S_k\}} & \quad C_{O}(a, \{S_k\}) \nonumber \\
\text{s.t.}           & \quad 0 \leq S_k \leq D_k, \forall k \in \mathcal{K}, \label{e:S_k_interval}\\
                         & \quad \sum_{k=1}^K S_k \geq S_T, \label{e:S_k_sum}\\
                         & \quad a>0.
\end{align}
\end{subequations}
\end{prob}
}

{
Comparing the output of $C_{I}(a, \{S_k\})$ and $C_{O}(a, \{S_k\})$ for any specific input of $a$ and $\{S_k\}$, there are two possible cases:
1) $S_k>0$ for every $k\in \mathcal{K}$; 2) there exist some $k \in \mathcal{K^{\dagger}} \neq \emptyset$ such that $S_k=0$.
\begin{itemize}
\item
For the first case, the cost function of the optimization problems associated with $C_{I}(a, \{S_k\})$ and $C_{O}(a, \{S_k\})$ (i.e., Problem \ref{p:raw_opt_orig_low}  and Problem \ref{p:raw_opt_low}) can be found to be identical since $S_k >0$ and $\mathcal{I}(S_k) = 1$ for every $k\in \mathcal{K}$, also Problem \ref{p:raw_opt_orig_low} and Problem \ref{p:raw_opt_low} have the same feasible region, so there is $C_{I}(a, \{S_k\}) = C_{O}(a, \{S_k\})$.
\item
For the second case, to achieve $C_{O}(a, \{S_k\})$, the optimal $b_k$ for $k\in \mathcal{K}^{\dagger}$ can be found to be zero so as to minimize the term $\left(a b_k h_k  - \left(\frac{S_k}{\sum_{k=1}^{K} S_k}\right) \right)^2 c_k $ in the cost function of Problem \ref{p:raw_opt_low}, since $S_k=0$ for $k\in \mathcal{K}^{\dagger}$. In this case, with $b_k$ replaced with its optimal solution (i.e., $b_k=0$) for $k\in \mathcal{K}^{\dagger}$, the cost function of Problem \ref{p:raw_opt_low} is exactly the cost function of Problem \ref{p:raw_opt_orig_low}.
Also Problem \ref{p:raw_opt_orig_low} and Problem \ref{p:raw_opt_low} have the same feasible region, then there is $C_{I}(a, \{S_k\}) = C_{O}(a, \{S_k\})$.
\end{itemize}
}

{
Summarizing the above discussion, we can state $C_{I}(a, \{S_k\})=C_{O}(a, \{S_k\})$ for any possible input of $a$ and $\{S_k\}$.
Recalling that Problem \ref{p:raw_opt_orig} is equivalent with Problem \ref{p:raw_opt_orig_up}, and Problem \ref{p:raw_opt} is equivalent with Problem \ref{p:raw_opt_up}, it is straightforward to see the equivalence between Problem \ref{p:raw_opt_orig} and Problem \ref{p:raw_opt}.
}

{
This completes the proof.
}

\section{Proof of Lemma \ref{lem:S2beta}} \label{app:S2beta}
For the mapping from the set of $\{S_k\}$ to the set of $\{\beta_k\}$, which is defined in (\ref{e:beta_def}), it can be easily checked that for any set of $\{S_k\}$ satisfying (\ref{e:S_k_interval}) and (\ref{e:S_k_sum}), the associated $\{\beta_k\}$, which are calculated according to (\ref{e:beta_def}), could always satisfy the constraints of (\ref{e:box_beta}) and (\ref{e:sum_beta}).
This proves the existence of the mapping from the set of $\{S_k\}$ restricted by (\ref{e:S_k_interval}) and (\ref{e:S_k_sum}) to the set of $\{\beta_k\}$ defined by (\ref{e:box_beta}) and (\ref{e:sum_beta}).

Next we need to prove the existence of the mapping from the set of $\{\beta_k\}$ satisfying (\ref{e:box_beta}) and (\ref{e:sum_beta}) to the set of $\{S_k\}$ defined by (\ref{e:S_k_interval}) and (\ref{e:S_k_sum}).
For a set of $\{\beta_k\}$ satisfying (\ref{e:box_beta}) and (\ref{e:sum_beta}), say $\tilde{\beta}_1$, $\tilde{\beta}_2$, ..., $\tilde{\beta}_K$,
define
\begin{equation} \label{e:Xi_exp_app}
\tilde{\Xi} = \max_{k \in \mathcal{K}} \frac{\tilde{\beta}_k}{D_k},
\end{equation}
where the optimal $k$ in (\ref{e:Xi_exp_app}) is denoted as $k^*$.
It can be found that the $\tilde{\Xi}$ given in (\ref{e:Xi_exp_app}) satisfies
\begin{equation}
\tilde{\Xi} = \frac{\tilde{\beta}_{k^*}}{D_{k^*}} \leq  \frac{\beta_{k^*}^{\max}}{D_{k^*}} = \frac{1}{S_T}.
\end{equation}
We then generate $\tilde{S}_k = \frac{\tilde{\beta}_k}{\tilde{\Xi}}$ for $k\in \mathcal{K}$, and we have
\begin{equation} \label{e:S_k_interval_app}
\tilde{S}_k = \tilde{\beta}_k/ \tilde{\Xi} \leq  \tilde{\beta}_k \frac{D_k} {\tilde{\beta}_k} \leq D_k, \forall k \in \mathcal{K},
\end{equation}
which holds since the defined $\tilde{\Xi}$ in (\ref{e:Xi_exp_app}) satisfies $\tilde{\Xi} \geq \frac{\tilde{\beta}_k}{D_k}$ for $\forall k \in \mathcal{K}$.
Additionally, it can be also found that
\begin{equation}  \label{e:S_k_sum_app}
\sum_{k=1}^{K} \tilde{S}_k  = \sum_{k=1}^{K} \frac{\tilde{\beta}_k}{\tilde{\Xi}} = \frac{1} {\tilde{\Xi}} \geq S_T.
\end{equation}
According to (\ref{e:S_k_interval_app}) and (\ref{e:S_k_sum_app}),
we can claim that a set of $\{\tilde{S}_k\}$ satisfying
(\ref{e:S_k_interval}) and (\ref{e:S_k_sum}) has been mapped from the set of $\{\tilde{\beta_k}\}$ satisfying (\ref{e:box_beta}) and (\ref{e:sum_beta}).

This completes the proof.

\section{Proof of Lemma \ref{lem:F_monotonicity}} \label{app:F_monotonicity}
In case that the set $\mathcal{K}_1(a)$ and the set $\mathcal{K}_2(a)$ keep unchanged, it can be checked that the function $P(a, x)$ is non-decreasing with both $a$ and $x$.
Hence for Problem \ref{p:F_a_def}, as $a$ grows, the feasible region of $x$ will not shrink (or will enlarge), which will lead to the nonincrease of its minimal cost function, i.e., $F(a)$.

This completes the proof.

\section{Proof of Lemma \ref{lem:F_convexity}} \label{app:F_convexity}
The proof is completed within two steps.
In the first step, we show that the function $P(a, x)$ is concave function with respect to $(a, x)^T$. This is because the term $\min \left(a h_k b_k^{\max} + x, \beta_k^{\max}\right)$ is a concave function with $(a, x)^T$ considering that it is the minimization of two linear functions with $(a, x)^T$, i.e., $\left(a h_k b_k^{\max} + x\right)$ and $\left( 0 \cdot a + 0 \cdot x + \beta_k^{\max}\right)$.

In the second step, suppose the optimal $x$ of Problem \ref{p:F_a_def} when  $a=a^{\dag}$ is $x^{\dag}$, and the optimal $x$ of Problem \ref{p:F_a_def} when $a=a^{\ddag}$ is $x^{\ddag}$. Then we have $P(a^{\dag}, x^{\dag}) \geq 1$ and $P(a^{\ddag}, x^{\ddag}) \geq 1$. For any $\theta \in [0, 1]$, we have
\begin{equation} \label{e:F_a_convexity_1}
\begin{split}
	1 = & ~\theta + \left(1 - \theta\right) \\
  \leq &  ~\theta P(a^{\dag}, x^{\dag})  + \left(1 - \theta\right) P(a^{\ddag}, x^{\dag}) \\
	\overset{(a)}{\leq} & ~ P\left(\theta a^{\dag} + (1-\theta) a^{\ddag}, \theta x^{\dag} + (1-\theta) x^{\ddag} \right)
	\end{split}
	\end{equation}
where $(a)$ holds since the function $P(a, x)$ is a concave function with $(a, x)^T$.
According to the statement of Problem \ref{p:F_a_def},  $x= \left(\theta x^{\dag} + (1-\theta) x^{\ddag}\right)$ is a feasible solution of Problem \ref{p:F_a_def} when $a = \left(\theta a^{\dag} + (1-\theta) a^{\ddag}\right)$, which is definitely no less than $F\left(\theta a^{\dag} + (1-\theta) a^{\ddag}\right)$. Then we have
\begin{equation} \label{e:F_a_convexity_2}
\begin{split}
&~\theta F(a^{\dag}) + (1-\theta) F(a^{\ddag}) \\
=& ~\theta x^{\dag} + (1-\theta) x^{\ddag} \\
\geq& ~ F\left(\theta a^{\dag} + (1-\theta) a^{\ddag}\right),
\end{split}
\end{equation}
which proves the convexity of function $F(a)$ with $a$.

This completes the proof.

%

\end{appendices}
\bibliography{references}
\bibliographystyle{IEEEtran}

\end{document}